\documentclass{article}

    \PassOptionsToPackage{numbers, compress}{natbib}

    \usepackage[preprint]{neurips_2025}

\usepackage[utf8]{inputenc} 
\usepackage[T1]{fontenc}    
\usepackage{hyperref}       
\usepackage{url}            
\usepackage{booktabs}       
\usepackage{amsfonts}       
\usepackage{nicefrac}       
\usepackage{microtype}      
\usepackage{xcolor}         
\usepackage{amsmath}
\usepackage{amsfonts} 
\usepackage{amssymb} 
\usepackage{multirow}  
\usepackage{graphicx}
\usepackage{float}
\usepackage{arydshln}
\usepackage{colortbl}  
\usepackage{array} 
\usepackage{algorithm}
\usepackage{algpseudocode} 
\usepackage{framed}

\title{An Empirical Analysis of VLM-based OOD Detection: Mechanisms, Advantages, and Sensitivity}

\author{%
Yuxiao Lee\textsuperscript{1}\\
\texttt{yuxiao9922@mails.jlu.edu.cn}
\And
Xiaofeng Cao\textsuperscript{2}\thanks{Corresponding author.}\\
\texttt{xiaofengcao@tongji.edu.cn}
\And
Wei Ye\textsuperscript{3} \\
\texttt{yew@tongji.edu.cn}
\And
Jiangchao Yao\textsuperscript{4}\\
\texttt{sunarker@sjtu.edu.cn}
\And
Jingkuan Song\textsuperscript{2}\\
\texttt{jingkuan.song@gmail.com}
\And
Heng Tao Shen\textsuperscript{2,6}\\
\texttt{shenhengtao@hotmail.com}
\And
\mdseries \textsuperscript{1}School of Artificial Intelligence, Jilin University \\
\textsuperscript{2} School of Computer Science and Technology, Tongji University  \\
\textsuperscript{3}College of Electronic and Information Engineering, Tongji University \\
\textsuperscript{4}CMIC, Shanghai Jiao Tong University \\
\textsuperscript{5}Engineering Research Center of Intelligent Finance, Ministry of Education \\
\textsuperscript{6}Center for Future Media, University of Electronic Science and Technology of China \\
}

\begin{document}

\maketitle


\begin{abstract}
Vision-Language Models (VLMs), such as CLIP, have demonstrated remarkable zero-shot out-of-distribution (OOD) detection capabilities, vital for reliable AI systems. Despite this promising capability, a comprehensive understanding of (1) \textbf{why} they work so effectively, (2) \textbf{what} advantages do they have over single-modal methods, and (3) \textbf{how} is their behavioral robustness — remains notably incomplete within the research community. This paper presents a systematic empirical analysis of VLM-based OOD detection using in-distribution (ID) and OOD prompts. (1) \textbf{Mechanisms:} We systematically characterize and formalize key operational properties within the VLM embedding space that facilitate zero-shot OOD detection.
(2) \textbf{Advantages:} We empirically quantify the superiority of these models over established single-modal approaches, attributing this distinct advantage to the VLM's capacity to leverage rich semantic novelty.
(3) \textbf{Sensitivity:} We uncovers a significant and previously under-explored asymmetry in their robustness profile: while exhibiting resilience to common image noise, these VLM-based methods are highly sensitive to prompt phrasing.
Our findings contribute a more structured understanding of the strengths and critical vulnerabilities inherent in VLM-based OOD detection, offering crucial, empirically-grounded guidance for developing more robust and reliable future designs.
\end{abstract}


\section{Introduction}
\label{sec:intro}

The remarkable capabilities of large pre-trained models, particularly Vision-Language Models (VLMs) like CLIP \citep{CLIP}, have revolutionized numerous downstream tasks, often enabling impressive zero-shot or few-shot performance without extensive task-specific fine-tuning. A key enabler of this versatility is their ability to map different modalities into a shared semantic embedding space, facilitating cross-modal understanding and transfer. While the empirical success of these models across various vision and language tasks is well-established \cite{chang2024survey,zhao2023survey}, systematically exploring and fully understanding the fundamental properties of their learned representations—and precisely how these properties enable or influence performance on challenging tasks like out-of-distribution (OOD) detection—remains a significant challenge. Indeed, a deeper, empirically-backed understanding of why these models excel and where their vulnerabilities lie is paramount for their trustworthy deployment.

Out-of-distribution detection is a critical problem for deploying machine learning systems reliably in open-world environments, demanding the ability to identify inputs that differ significantly from the training data distribution \cite{survey1,survey2}. Compared to traditional OOD detection methods, which primarily rely on single-modal data statistics or features from models trained for specific tasks \cite{hendrycks2016baseline}, approaches leveraging large-scale pre-trained VLMs have demonstrated superior performance and generalizability in detecting semantic novelty \cite{ramesh2022hierarchical,wang2022clip,crowson2022vqgan,wang2021actionclip,gao2024clip}. These VLM-based methods typically utilize textual prompts to assess the similarity between image embeddings and prompt embeddings within the VLM's joint space. Despite their empirical effectiveness, a granular understanding of the precise mechanisms within the VLM's embedding space that facilitate OOD detection, their advantages over single-modal baselines, and their specific behavioral characteristics under varied conditions are not yet systematically understood or quantitatively detailed. This lack of in-depth, structured understanding poses a significant barrier to developing truly trustworthy, interpretable, and robust VLM-based OOD detection systems. Addressing this analytical gap by providing a rigorous empirical foundation is therefore crucial for designing more robust and theoretically grounded OOD detection systems.

\begin{figure*}[h!]
\begin{center}
\includegraphics[width=0.85\textwidth]{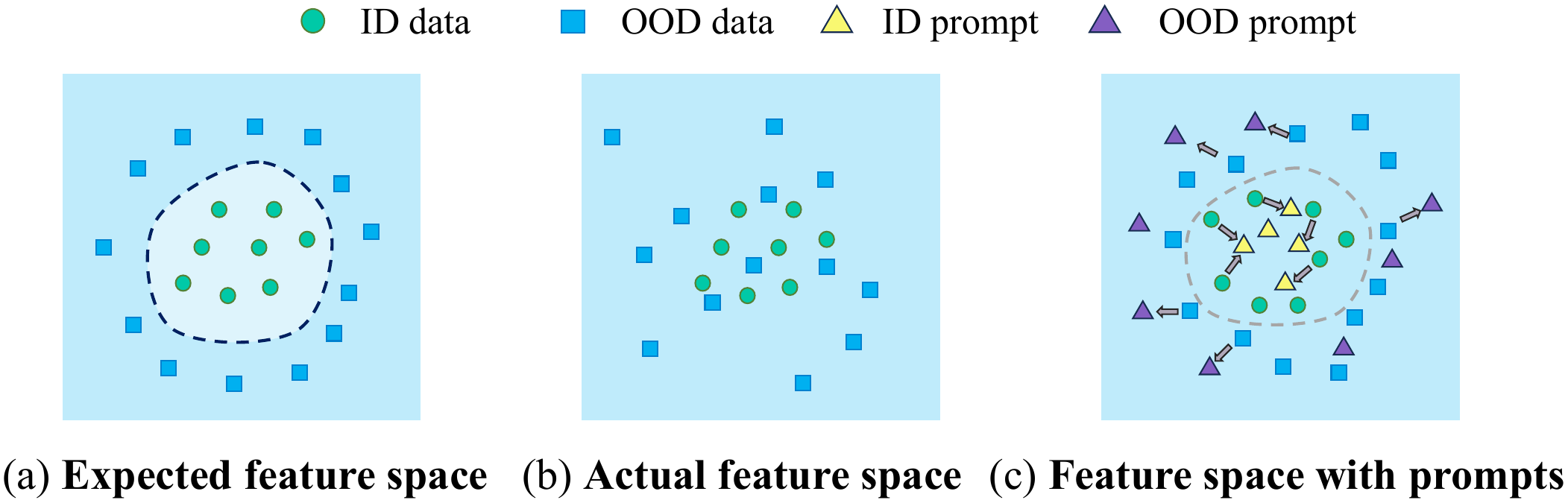}
\end{center}
  \vspace{-.05in}
  \caption{(a) \textbf{Expected feature space} illustrating how ID data is hypothesized to be well-separated from OOD data. (b) \textbf{Actual feature space} showing less distinct separation between ID and OOD data in practice. (c) \textbf{Feature space with prompts}, where ID and OOD prompts assist in organizing the data, thereby enhancing the separation between ID and OOD data points.
  }
  \label{fig1: motivation}
\vspace{-.05in}
\end{figure*}

To bridge this critical analytical gap and offer a clearer, evidence-based perspective on VLM-based OOD detection, this paper presents a systematic empirical analysis. We aim to dissect its core operational principles, quantify its advantages with comprehensive comparisons, and characterize its behavioral nuances under various perturbations. Our analysis focuses on a prevalent paradigm that incorporates both in-distribution (ID) and explicitly designed out-of-distribution (OOD) prompts—a representative and effective framework for probing the VLM's semantic space (as conceptually depicted, for instance, in Figure~\ref{fig: framework}). As conceptually illustrated in Figure~\ref{fig1: motivation}, such methods exploit the learned semantic structure of the VLM space, guided by textual prompts representing ID and OOD concepts, to achieve improved separation. The core idea is to compute a score that reflects an image's affinity to known ID concepts relative to other relevant concepts (including OOD ones). While the application of these methods has shown promise, our work moves beyond mere application or incremental improvement; we aim to provide a deeper, data-driven understanding of the fundamental empirically observable properties of the VLM space that underpin their OOD detection performance and how these systems behave under different conditions. This foundational understanding is essential for guiding future research and development.

Building upon extensive empirical observations, our analysis reveals several key insights into VLM-based OOD detection:
First, addressing the fundamental question of why the VLM's joint embedding space enables zero-shot OOD detection, we present a systematic empirical characterization of the underlying mechanisms. This involves formalizing and providing robust quantitative evidence for key observed properties such as ID classification alignment and the enhanced separability achieved through relative affinity to ID and OOD prompts. 
Second, to understand the advantages that VLM-based methods empirically outperform single-modal approaches, we conduct a rigorous empirical analysis and explain the reasons behind this observed superiority. We provide extensive comparative results highlighting the VLM's proficiency in leveraging its rich, cross-modally learned semantic space for detecting conceptual novelty—a capability often beyond the reach of unimodal systems.
Third, recognizing that practical deployment necessitates robustness, and given the inherent multi-modal nature of VLM inputs, we investigate their behavior under perturbations across modalities. We conduct a detailed study characterizing the robustness and sensitivity of VLM-based OOD detection to image and text perturbations. This uncovers a significant and potentially critical asymmetry: a notable resilience to image noise contrasted with a high sensitivity to prompt phrasing and a striking vulnerability to textual perturbations, highlighting a key area for future improvement.

The contributions of this work are therefore summarized as follows:
\begin{itemize}
\item We provide a systematic empirical characterization and formalization of the fundamental mechanisms within the VLM joint embedding space that enable zero-shot OOD detection using ID and OOD prompts. This offers a clear, evidence-based account of how these systems operate.
\item We conduct a comprehensive empirical analysis that quantitatively demonstrates and explains the superiority of VLM-based OOD detection over single-modal image-based methods, emphasizing the VLM's crucial advantage in leveraging its learned semantic space to identify conceptual novelty.
\item We present a detailed study of the robustness and sensitivity of VLM-based OOD detection, revealing a critical asymmetric vulnerability to textual perturbations compared to image perturbations. This finding has significant implications for the reliable deployment of such systems.
\end{itemize}

\section{Preliminaries}
\label{sec:pre}


\paragraph{Out-of-Distribution Detection Problem}
The task of Out-of-Distribution (OOD) detection is fundamental to deploying machine learning models reliably in open-world scenarios. It focuses on identifying whether an input sample originates from the same data distribution as the training data (in-distribution, ID) or from a different distribution (out-of-distribution, OOD). Formally, let $\mathcal{X}$ be the input space, and let $\mathcal{D}_{\tiny\text{ID}}$ and $\mathcal{D}_{\text{OOD}}$ denote the probability distributions of ID and OOD data, respectively, where samples from $\mathcal{D}_{\tiny\text{OOD}}$ are distinct from $\mathcal{D}_{\tiny\text{ID}}$. The objective is to devise a function or mechanism that can effectively discriminate between a sample $x \sim \mathcal{D}_{\tiny\text{ID}}$ and a sample $x' \sim \mathcal{D}_{\tiny\text{OOD}}$. Typically, OOD detection methods involve computing an anomaly or uncertainty score $S(x)$ for a given input $x$. A predetermined threshold $\lambda$ is then applied to this score: if $S(x)$ exceeds $\lambda$, the sample is classified as ID; otherwise, it is classified as OOD. This problem is crucial for safety-critical applications where encountering novel or potentially harmful inputs must be detected.

\paragraph{Contrastive Vision-Language Models}
Our analysis is centered around powerful pre-trained contrastive vision-language models (VLMs), such as CLIP, which learn to align image and text representations in a shared embedding space through a contrastive training objective on large-scale multimodal datasets. A VLM typically comprises an image encoder $\phi_I: \mathcal{X} \to \mathcal{E}$ and a text encoder $\phi_T: \mathcal{T} \to \mathcal{E}$, where $\mathcal{T}$ is the text space and $\mathcal{E} \cong \mathbb{R}^d$ is the $d$-dimensional joint semantic embedding space. The core idea behind contrastive training is to maximize the similarity (e.g., cosine similarity, denoted $\mathrm{sim}(\cdot,\cdot)$) between the embeddings of truly paired images and texts, while simultaneously minimizing the similarity between embeddings of incorrectly paired samples. This process enables the VLM to capture rich visual and semantic concepts and their associations, facilitating remarkable zero-shot capabilities on various downstream tasks without requiring task-specific fine-tuning. Architectures like the Vision Transformer (ViT) are often employed as the image encoder, allowing images to be processed similarly to sequences of text tokens.

\begin{figure*}[h]
\begin{center}
\includegraphics[width=0.90\textwidth]{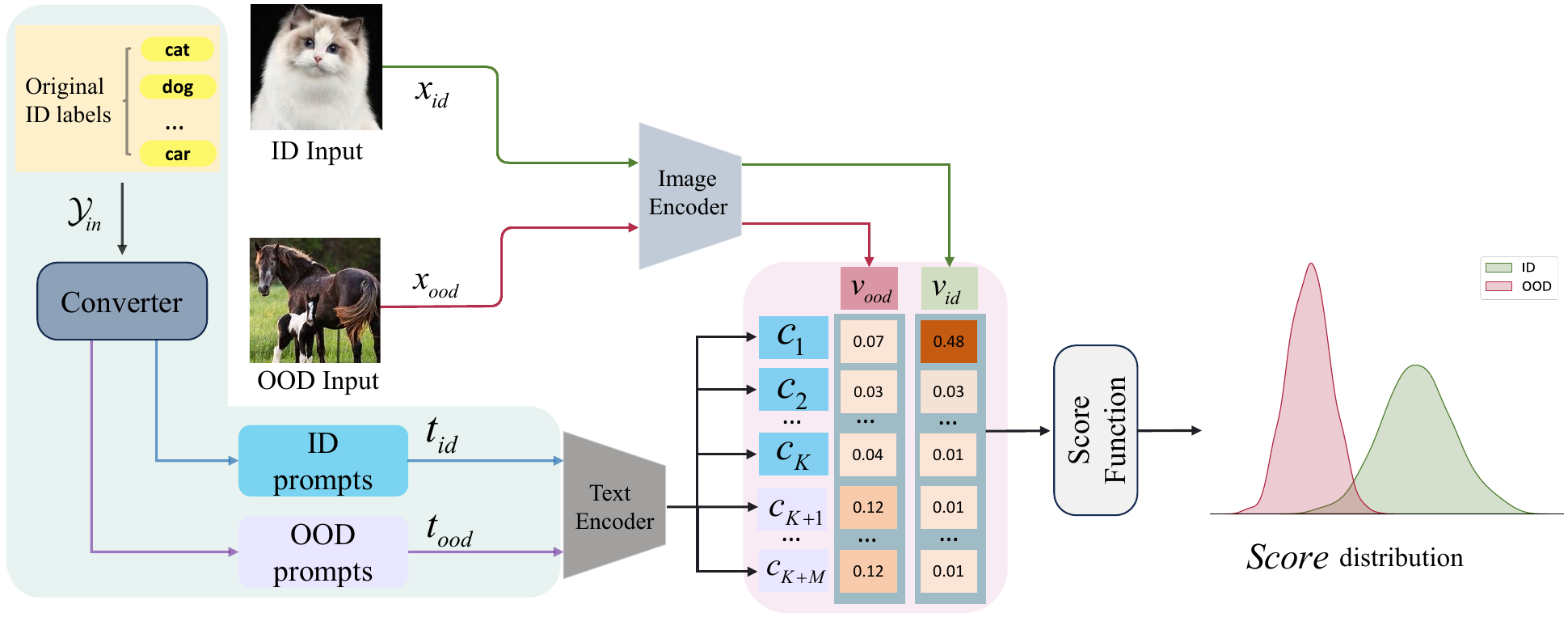}
\end{center}
  \vspace{-.05in}
  \caption{A unified OOD detection pipeline leveraging VLMs with ID and OOD prompts. Original ID class labels are transformed into a set of ID prompts (e.g., ``\texttt{a photo of a \{label\}}'') and a set of predefined OOD prompts (e.g., ``\texttt{an unrelated object}'') by a \textit{Converter} module. An input image $x_i$ (either ID or OOD) is encoded by the image encoder $g(\cdot)$ into a visual embedding $\mathbf{v}$. Similarly, ID prompts $t_{id}$ and OOD prompts $t_{ood}$ are encoded by the text encoder $f(\cdot)$ into text embeddings $\mathbf{c}_{id}$ and $\mathbf{c}_{ood}$. Similarity scores (e.g., cosine similarity) between $\mathbf{v}$ and all prompt embeddings are computed. These scores are then aggregated, often using a normalized function (as in Eq.~\eqref{eq:score_id_ood}), to produce a final OOD score, whose distribution helps discriminate between ID and OOD inputs. Further details are provided in Appendix~\ref{appendix:subsec:algorithm}.
  }
   \label{fig: framework}
  \vspace{-.05in}
\end{figure*}

\paragraph{VLM-based OOD Detection with ID and OOD Prompts}
Vision-language models have emerged as powerful tools for zero-shot out-of-distribution detection, leveraging their extensive pre-training on diverse image-text data. A common paradigm involves mapping both input images and textual descriptions of known in-distribution categories into a shared semantic embedding space $\mathbb{R}^d$. OOD detection is then performed by assessing the similarity between the image embedding and these ID concept embeddings (or "prototypes"). While many initial approaches rely solely on ID prompts, our analysis focuses on a prevalent extension that incorporates both ID and dedicated OOD prompts to enhance detection capabilities. As conceptually illustrated in Figure~\ref{fig: framework}, this approach utilizes the VLM's image encoder $g(\cdot)$ to obtain a visual feature $\mathbf{v} \in \mathbb{R}^d$ for an input image $x$, and its text encoder $f(\cdot)$ to generate embeddings for a set of $K$ ID category prompts $\{\mathbf{c}_1, \dots, \mathbf{c}_K\}$ and $M$ carefully constructed OOD prompts $\{\mathbf{c}_{K+1}, \dots, \mathbf{c}_{K+M}\}$. ID prompts are typically formed by applying templates (e.g., ``\texttt{a photo of a \{label\}}'') to the known ID class labels. In contrast, OOD prompts are designed to represent abstract, neutral, or unrelated concepts (e.g., ``\texttt{an unrelated object}'') that are expected to have higher affinity with OOD images than ID images. The core mechanism involves computing a unified score that quantifies the image's affinity to the set of ID concepts relative to its affinity to the combined set of ID and OOD concepts. Specifically, this score is often computed by taking the similarity of the image embedding $\mathbf{v}$ to the best-matching ID prototype $\mathbf{c}_{\hat{k}}$ ($\hat{k} = \arg\max_{1 \leq i \leq K} \mathrm{sim}(\mathbf{v}, \mathbf{c}_i)$) and normalizing it by the sum of exponentiated similarities to all $K+M$ prototypes. For instance, a common scoring function analogous to Equation \eqref{eq:score_id_ood} yields a high score for images strongly aligning with an ID prototype but weakly with OOD prototypes, characteristic of ID samples. Conversely, images with weaker ID alignment or stronger OOD alignment will result in a lower score, indicative of OOD samples. The complete step-by-step process for this type of VLM-based OOD detection is typically outlined in an algorithm, such as Algorithm~\ref{alg:Unified_OOD_Detection_Pipeline}. It is this specific pipeline, leveraging the interplay between ID and OOD prompt affinities within the VLM's embedding space, that forms the basis of our subsequent empirical analysis.

\section{Mechanisms of VLM-based OOD Detection}
\label{sec:mechanisms}
\vspace{-.05in}
\begin{center}
\fcolorbox{black}{yellow!10}{\parbox{0.98\linewidth}{\textit{\textbf{Question: Why VLM-based OOD Detection work so effectively?}}}}
\end{center}
\vspace{-.05in}

Vision-language models like CLIP have demonstrated remarkable zero-shot OOD detection capabilities. This proficiency is not accidental but is rooted in the inherent structure of their learned joint image-text embedding space $\mathcal{E} \cong \mathbb{R}^d$. Unlike traditional methods that often rely on features optimized for single-modal classification or density estimation on ID data, VLMs, through large-scale contrastive pre-training, learn a semantic space where similarity reflects conceptual relatedness between images and text. This section dissects and provides systematic empirical validation for the fundamental, observable properties of this space that collectively explain how VLM-based approaches, particularly those utilizing both ID and OOD prompts, effectively distinguish between in-distribution and out-of-distribution samples. While the high-level principles of contrastive learning are known, their specific manifestations and interplay in the context of prompt-based OOD detection warrant detailed empirical characterization. Our analysis formalizes these key operational properties based on extensive empirical observations within $\mathcal{E}$, as summarized in Insight~\hyperref[insight1]{1} below. This formalization aims to provide a clear framework for understanding these mechanisms, grounded in quantifiable evidence.

\textbf{Insight 1}\label{insight1} 
\emph{Let $\mathcal{I}$ and $\mathcal{T}$ be the image and text spaces, respectively, and $\mathcal{E} \cong \mathbb{R}^d$ the shared semantic embedding space with vision-language model (VLM) embedding functions $\phi_I, \phi_T$. The similarity function is denoted by $f(\mathbf{u}, \mathbf{v})$. Let $\mathcal{D}_\mathrm{ID}$ be the in-distribution data distribution over $N$ classes $C_\mathrm{ID}$, with specific prompts $\{P_c\}_{c \in C_\mathrm{ID}}$ for each class $c$, and $P_\mathrm{OOD}$ an OOD prompt. Define $S_\mathrm{ID}(I) = \max_{c \in C_\mathrm{ID}} s(\phi_I(I), \phi_T(P_c))$ and $S_\mathrm{OOD}(I) = f(\phi_I(I), \phi_T(P_\mathrm{OOD}))$. The mechanism of VLM-based OOD detection relies on the following empirically observed properties within $\mathcal{E}$:}

1) \emph{For an ID image $I_\mathrm{ID} \sim \mathcal{D}_\mathrm{ID}$ from class $c^*$, its embedding is, on average, maximally similar to the embedding of the true class prompt among all ID prompts:}
    \begin{equation}
    \label{property1-insight1}
    \mathbb{E}_{I_\mathrm{ID} \sim \mathcal{D}_\mathrm{ID}}[f(\phi_I(I_\mathrm{ID}), \phi_T(P_{c^*}))] > \mathbb{E}_{I_\mathrm{ID} \sim \mathcal{D}_\mathrm{ID}}[\max_{c \in C_\mathrm{ID}, c \neq c^*} f(\phi_I(I_\mathrm{ID}), \phi_T(P_c))];
    \end{equation}
2) \emph{The maximum similarity to ID class prompts, $S_\mathrm{ID}(I)$, tends to be higher for ID samples than for OOD samples:}
    \begin{equation}
    \label{property2-insight1}
    \mathbb{E}_{I_\mathrm{ID} \sim \mathcal{D}_\mathrm{ID}}[S_\mathrm{ID}(I_\mathrm{ID})] > \mathbb{E}_{I_\mathrm{OOD} \notin \mathcal{D}_\mathrm{ID}}[S_\mathrm{ID}(I_\mathrm{OOD})]; 
    \end{equation}
    \emph{Specifically, empirical findings suggest the existence of a threshold $\tau_\mathrm{ID}$ such that for many $I_\mathrm{OOD} \notin \mathcal{D}_\mathrm{ID}$, $S_\mathrm{ID}(I_\mathrm{OOD}) \leq \tau_\mathrm{ID}$.}
    
3) \emph{The difference score $S_\mathrm{ID}(I) - S_\mathrm{OOD}(I)$ provides improved separation between ID and OOD samples, due to its higher expected value for ID images:}
    \begin{equation}
    \label{property3-insight1}
    \mathbb{E}_{I_\mathrm{ID} \sim \mathcal{D}_\mathrm{ID}}[S_\mathrm{ID}(I_\mathrm{ID}) - S_\mathrm{OOD}(I_\mathrm{ID})] > \mathbb{E}_{I_\mathrm{OOD} \notin \mathcal{D}_\mathrm{ID}}[S_\mathrm{ID}(I_\mathrm{OOD}) - S_\mathrm{OOD}(I_\mathrm{OOD})], 
    \end{equation}
    \emph{where the expectation for $I_\mathrm{OOD}$ is taken over a relevant distribution of out-of-distribution data.}
    
This Insight formalizes fundamental empirical observations, systematically cataloged in our study, about the structure of the VLM joint image-text embedding space and how these specific, quantifiable properties enable prompt-based out-of-distribution detection. The core principle lies in how the VLM's contrastive training objective shapes the embedding space $\mathcal{E}$ by forcing embeddings of semantically related image-text pairs closer while pushing unrelated pairs apart, thus learning a shared space where cosine similarity measures semantic relatedness. While this general training principle is understood, our contribution lies in the detailed empirical characterization of its consequences for OOD detection when guided by explicit ID and OOD prompts. In the following subsections, we empirically demonstrate and elaborate on each of these properties, supported by visualizations (e.g., Figure~\ref{fig: mechanisms_visualization}) and quantitative results detailed in the Appendices. Our aim is to provide a clear, evidence-based narrative of these operational mechanisms.

\begin{figure*}[h]
\begin{center}
\includegraphics[width=0.85\textwidth]{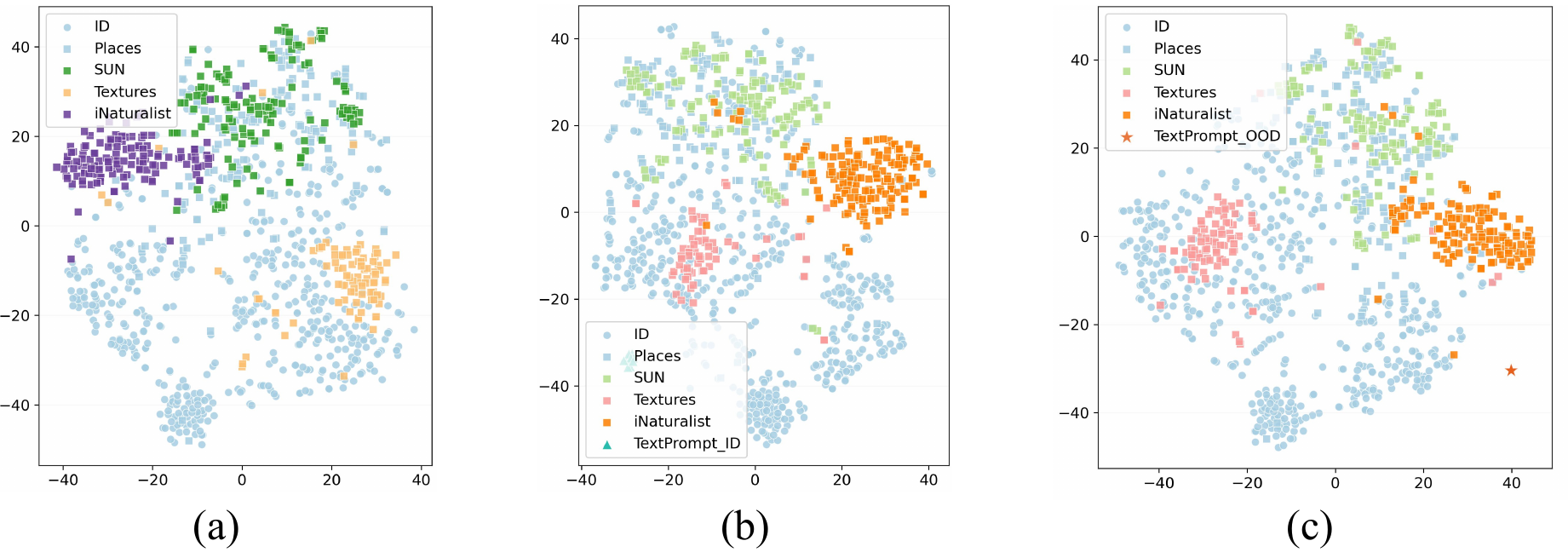}
\end{center}
    \caption{Visualizations of the VLM embedding space illustrating key properties for OOD detection (Insight~\ref{insight1}). (a) A 2D projection of image embeddings for ID data (from ImageNet-1k) and various OOD datasets, showing their relative spatial arrangement. (b) A 2D projection showing ID image embeddings and their corresponding ID class prompt embeddings (TextPrompt\_ID), demonstrating the alignment between images and true class concepts. A noteworthy observation is that, in the projection, the Textures dataset clusters closely with the ID data, which corresponds to the model's poor performance in distinguishing ID from Textures datasets. (c) A 2D projection showing ID and OOD image embeddings along with an embedding of a representative OOD concept prompt (TextPrompt\_OOD), illustrating how OOD prompts position themselves relative to ID and OOD image clusters. These visualizations provide qualitative support for the empirically observed properties discussed in Section~\ref{sec:mechanisms}.}
    \label{fig: mechanisms_visualization}
  \vspace{-.05in}
\end{figure*}

\subsection{ID Classification Alignment}
\label{subsec:ID classification}
The first fundamental property enabling VLM-based OOD detection is the inherent ability of VLMs to align image embeddings with the text embeddings of their corresponding semantic concepts within the learned space $\mathcal{E}$. As formalized in Insight~\hyperref[insight1]{1} (\ref{property1-insight1}), for an image $I_\mathrm{ID}$ truly belonging to class $c^*$, its embedding $\phi_I(I_\mathrm{ID})$ is expected, on average, to have the highest similarity with the true class prompt embedding $\phi_T(P_{c^*})$ compared to any other ID class prompt embedding $\phi_T(P_c)$ for $c \neq c^*$.

This property is a direct consequence of the VLM's contrastive training objective. By training on massive datasets of image-text pairs, the model learns to bring embeddings of matched pairs (like an image of a cat and the text "a photo of a cat") closer in the embedding space, while pushing embeddings of mismatched pairs apart. For ID samples, the prompt corresponding to the true class serves as the matched text, leading to high similarity. To validate this property, we analyze the similarity scores between ID image embeddings and all ID class prompt embeddings. The complete results are shown in Appendix~\ref{appendix:subsubsec:id_classification_alignment_appendix}. This internal classification capability forms the basis for calculating $S_\mathrm{ID}(I)$, the maximum similarity to any ID prompt, which is subsequently used for OOD detection.

\subsection{ID vs. OOD Maximum ID Similarity Contrast}
\label{subsec:IDvsOOD}
Building upon the ID Classification Alignment, the second critical property is the contrast in the maximum similarity to ID prompts between ID and OOD samples. Insight~\hyperref[insight1]{1} (\ref{property2-insight1}) states that the maximum similarity to any ID class prompt, $S_\mathrm{ID}(I)$, is expected to be significantly higher for ID samples $I_\mathrm{ID}$ than for OOD samples $I_\mathrm{OOD}$.

The intuition behind this property is that OOD images represent concepts or visual patterns that are distinct from the set of known ID categories used to define the ID prompts. While an ID image will typically align strongly with its true class prompt (Property~\ref{property1-insight1}), an OOD image is less likely to exhibit a strong semantic match with any of the predefined ID class prompts. Consequently, the maximum similarity score $S_\mathrm{ID}(I_\mathrm{OOD})$ for an OOD image is generally lower than $S_\mathrm{ID}(I_\mathrm{ID})$ for an ID image. This difference in expected maximum similarity creates a separability between the two distributions. We empirical validate this in Appendix~\ref{appendix:subsubsec:id_vs_ood_max_sim_appendix}. This property is the core mechanism exploited by VLM-based methods that rely solely on maximum ID similarity for discrimination.

\subsection{Relative Affinity Separation}
\label{subsec:separation}
While the contrast in $S_\mathrm{ID}(I)$ provides a baseline for OOD detection, many effective VLM-based methods leverage additional information, notably from explicitly designed OOD concept prompts. This leads to the third property, the Relative Affinity Separation, formalized in Insight~\hyperref[insight1]{1} (\ref{property3-insight1}). This property asserts that scores incorporating the relative affinity to ID versus OOD concepts, such as the difference score like the one presented in Equation \eqref{eq:score_id_ood}), provide improved separation between ID and OOD samples compared to using $S_\mathrm{ID}(I)$ alone.

\begin{figure*}[h]
\begin{center}
\includegraphics[width=0.80\textwidth]{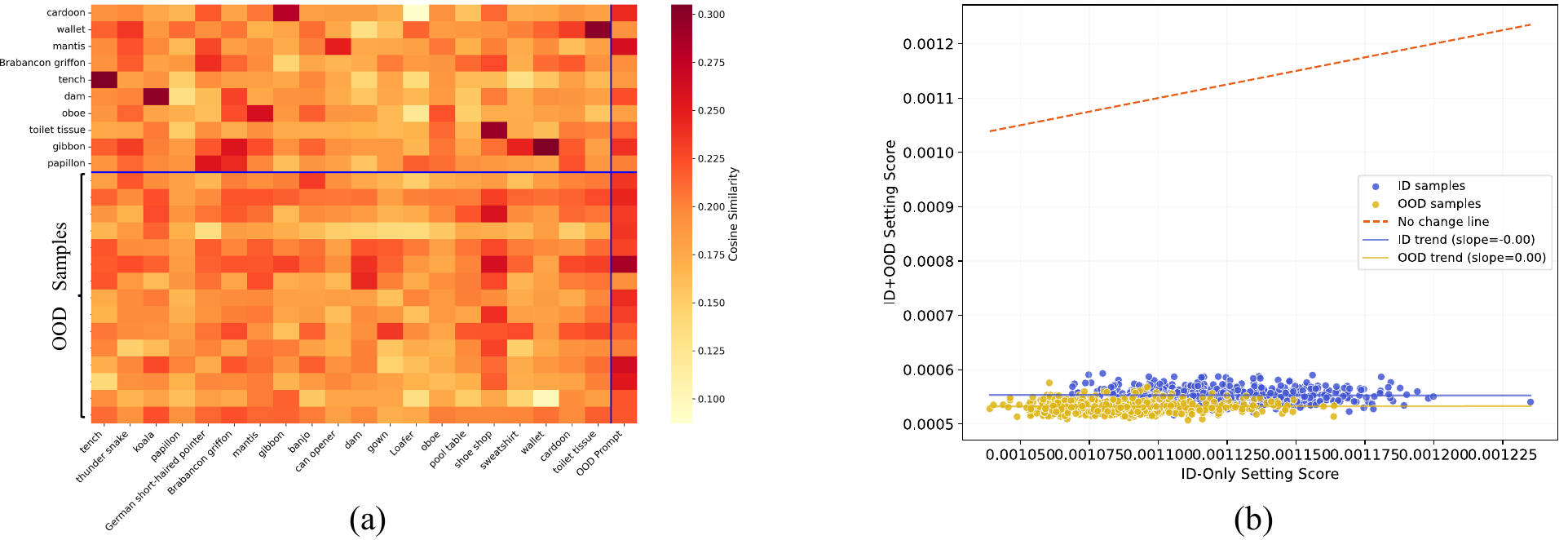}
\end{center}
    \caption{Empirical evidence supporting the Relative Affinity Separation Property. (a) Cosine similarity heatmap between embeddings of example ID images and OOD images and embeddings of various ID class prompts and one representative OOD prompt. The heatmap illustrates how OOD samples may exhibit varying degrees of similarity to ID concepts but potentially distinct affinities to the OOD prompt. (b) Scatter plot showing the distribution of scores for ID and OOD samples using an ID-only score ($S_{\text{ID}}(x)$) versus a score incorporating OOD prompts ($S_{\text{ID+OOD}}(x)$). The plot qualitatively demonstrates that the score leveraging relative affinities to both ID and OOD prompts provides better separation between ID and OOD samples compared to the ID-only score.}
    \label{fig: relative_affinity_separation} 

  \vspace{-.05in}
\end{figure*}

The rationale is that incorporating $S_\mathrm{OOD}(I)$ (similarity to an OOD prompt) provides a more nuanced view of an image's position in the semantic space relative to both known ID concepts and explicitly defined OOD concepts. For an ID image, $S_\mathrm{ID}(I)$ is high (Property~\ref{property1-insight1} \& \ref{property2-insight1}), and $S_\mathrm{OOD}(I)$ is expected to be relatively low. The difference will thus be high. For an OOD image, $S_\mathrm{ID}(I)$ is typically lower than for ID (Property~\ref{property2-insight1}). Furthermore, a well-chosen OOD prompt $P_\mathrm{OOD}$ might capture aspects of OOD data, leading to a higher $S_\mathrm{OOD}(I)$ for OOD images compared to ID images. The difference score then potentially becomes lower for OOD samples, amplifying the separation signal beyond what $S_\mathrm{ID}(I)$ alone provides. This mechanism is precisely what methods utilizing both ID and OOD prompts (like the one described in Section~\ref{sec:pre}) exploit. We empirically investigate the effectiveness of scores based on relative affinity in Appendix ~\ref{appendix:subsubsec:relative_affinity_scoring_appendix}, such as the unified score $S_{\mathrm{ID+OOD}}(x)$ from Equation \eqref{eq:score_id_ood}, in separating ID from OOD samples. Figure~\ref{fig: relative_affinity_separation} provides the most intuitive visualization result.

\vspace{-.05in}
\begin{center}
\fcolorbox{black}{green!10}{\parbox{0.98\linewidth}{\textit{\textbf{Takeaway:} VLM-based OOD detection effectively works by leveraging its learned semantic space for prompt-guided ID/OOD differentiation, further amplified by dedicated OOD prompts.}}}
\end{center}
\vspace{-.05in}

\section{Advantage over Single-Modal Methods}
\label{sec:advantage-single-modal}
\vspace{-.05in}
\begin{center}
\fcolorbox{black}{yellow!10}{\parbox{0.98\linewidth}{\textit{\textbf{Question: What advantages do VLM-based OOD detection have over single-modal methods?}}}}
\end{center}
\vspace{-.05in}

While traditional out-of-distribution detection methods have made significant progress, they often operate primarily on single-modal data, typically focusing on the statistical properties of image features learned during training on in-distribution data. The advent of powerful vision-language models has introduced a new paradigm, demonstrating empirically superior performance in detecting various types of OOD samples, particularly those exhibiting semantic novelty rather than just low-level visual shifts. This section presents our empirical analysis demonstrating this performance advantage and discusses the underlying reasons, grounded in the VLM's ability to leverage a rich semantic space. Our findings are encapsulated in Insight \hyperref[insight2]{2}:

\textbf{Insight 2}\label{insight2}
\emph{Let $S$ denote an out-of-distribution detection system or approach. Let $D_\mathrm{OOD}(S)$ be a scalar measure of its semantic image OOD detection performance, where higher values indicate better performance. We define and compare the following systems:} \emph{(a) $S_{\mathrm{VLM}}$: An OOD detection system based on a Vision-Language Model. This system utilizes both an image encoder ($\mathcal{M}_I$) and a text encoder ($\mathcal{M}_T$).} \emph{(b) $S_{\mathrm{SM},I}$: A single-modal OOD detection system relying solely on an image encoder (e.g., $\mathcal{M}_I$) and image-derived features, without leveraging textual prompts in the same manner as $S_{\mathrm{VLM}}$.} \emph{(c) $S_{\mathrm{SM},T}$: A hypothetical single-modal system for image OOD detection that would rely solely on a text encoder ($\mathcal{M}_T$).} \emph{Based on our systematic empirical observations, the following relationship generally holds regarding their OOD detection performance on image inputs:}
\begin{equation}
\label{property-insight2}
D_{\scriptscriptstyle\mathrm{OOD}}(S_{\scriptscriptstyle\mathrm{VLM}}) \geq \min\{D_{\scriptscriptstyle\mathrm{OOD}}(S_{\scriptscriptstyle\mathrm{SM},I}), D_{\scriptscriptstyle\mathrm{OOD}}(S_{\scriptscriptstyle\mathrm{SM},T})\}.
\end{equation}
This insight states that the semantic image OOD detection performance achieved by a cross-modal VLM method is empirically observed to be at least as good as the minimum performance attainable using only representations from its unimodal components independently. Given the negligible information about image typically provided by a single-modal text representation alone ($D_\mathrm{OOD}(S_{\mathrm{SM},T}) \approx 0$ for image OOD detection), this empirically implies that cross-modal VLM approaches generally achieve performance comparable to, and typically significantly exceeding, that of methods relying solely on single-modal image representations ($D_\mathrm{OOD}(S_{\mathrm{VLM}}) \geq D_\mathrm{OOD}(S_{\mathrm{SM},I})$). The consistent empirical superiority of VLM-based methods arises from their ability to effectively utilize a rich, cross-modal semantic embedding space for detecting conceptual novelty, a capability that fundamentally surpasses the limits of methods focused purely on low-level image statistics or feature distributions learned from single-modal ID data. We elaborate on this through empirical comparison and subsequent discussion.

\subsection{Performance Comparison}
\label{subsec:performace}
To quantitatively assess the advantage of VLM-based OOD detection, we conduct a comprehensive empirical comparison against representative single-modal image-based OOD detection methods. The objective is to demonstrate that VLM methods consistently achieve superior performance across various ID and OOD dataset combinations. Appendix~\ref{appendix:subsec:detailed performance evaluation} provides the complete experimental procedures and results. These results empirically demonstrate the relationship suggested by Insight~\hyperref[insight2]{2}, showing that VLM performance not only meets but typically exceeds the performance of single-modal image methods.

\subsection{Discussion: Leveraging Semantic Space}
\label{subsec:space}

Single-modal image models, even large pre-trained ones, primarily learn representations optimized for tasks defined solely within the image domain, such as image classification. OOD detection methods built upon these features often rely on detecting deviations from the statistical distribution of the ID data in the feature space. They are effective at identifying samples that are visually anomalous or fall outside the learned manifold of ID images. However, they may struggle with OOD samples that are visually similar to ID data but represent entirely different semantic concepts.

In contrast, VLMs, through contrastive training on vast quantities of image-text pairs, learn a shared embedding space $\mathcal{E}$ that captures deep semantic relationships between visual and textual concepts. As discussed in Section~\ref{sec:mechanisms}, this space exhibits properties like ID classification alignment and relative affinity separation. By using textual prompts to define the semantic boundaries of known ID concepts and also introducing prompts for OOD concepts, VLM-based methods can assess an image's position within this semantic map. An OOD image might still be visually plausible (thus not strongly deviating in a single-modal image feature space), but it will likely exhibit a different pattern of affinities to the ID vs. OOD concept prompts in the VLM's semantic space compared to an ID image. For example, an image of a "chair" (ID) will have high similarity to "a photo of a chair" and low similarity to "an unrelated object," while an image of a "car" (OOD) might have lower similarity to "a photo of a chair" and potentially higher similarity to "an unrelated object" or other general concepts. This ability to evaluate semantic coherence with diverse concepts defined by text is the key advantage.

Therefore, the empirical superiority of VLM-based OOD detection methods arises from their ability to transcend purely image-centric statistical anomaly detection and leverage the rich, cross-modal semantic information embedded in the VLM's representation space. The use of text prompts provides a flexible and powerful mechanism to probe this semantic space and establish criteria for OOD detection based on conceptual novelty, which is particularly effective against semantic OOD shifts. This explains why $D_{\scriptscriptstyle\mathrm{OOD}}(S_{\scriptscriptstyle\mathrm{VLM}})$ is typically much higher than $D_\mathrm{OOD}(S_{\mathrm{SM},I})$ in practice.

\vspace{-.05in}
\begin{center}
\fcolorbox{black}{green!10}{\parbox{0.98\linewidth}{\textit{\textbf{Takeaway: } VLM-based OOD detection surpasses single-modal methods by using its rich semantic space and textual guidance to detect conceptual novelty beyond purely visual analysis.}}}
\end{center}
\vspace{-.05in}

\section{Robustness and Sensitivity Analysis}
\label{sec:analysis}

\vspace{-.05in}
\begin{center}
\fcolorbox{black}{yellow!10}{\parbox{0.98\linewidth}{\textit{\textbf{Question: How is VLM-based OOD detection's behavioral robustness?}}}}
\end{center}
\vspace{-.05in}

The deployment of out-of-distribution detection systems in real-world applications necessitates a thorough understanding of their behavior under various conditions, particularly their resilience to input variations and perturbations. For VLM-based OOD detection, which inherently processes information from two modalities (image and text), it is crucial to investigate robustness and sensitivity across both domains. This section presents a systematic empirical analysis of how different types of perturbations applied to images and textual prompts affect the performance of VLM-based OOD detection. Our findings, which constitute a significant contribution of this work, reveal a notable and somewhat surprising asymmetry in robustness between the modalities, highlighting important considerations for the reliability and security of these methods. These observations are summarized in Insight~\hyperref[insight3]{3}:

\textbf{Insight 3}\label{insight3}
\emph{Let $M$ be a VLM-based Out-of-Distribution detection method, and $\mathcal{H}(M, I, P)$ its performance measure for image $I$ and prompt $P$. Based on empirical observations, the following properties are commonly observed for typical inputs:}

1) \emph{For a typical image $I$, prompt $P$, and bounded image noise $\eta_I$ such that $I + \eta_I$ remains within the image domain, the performance exhibits robustness to image noise:}
\begin{equation}
\label{property1-insight3}
\mathcal{H}(M, I + \eta_I, P) \approx \mathcal{H}(M, I, P);
\end{equation}
2) \emph{For a typical image $I$ and a baseline prompt set $P_1$, let $P_2$ be a prompt set drawn from a distribution $\mathcal{D}_{\text{var}}(P_1)$ representing significant variations in wording or structure from $P_1$ while both are intended for the same OOD task, the performance exhibits sensitivity to prompt wording variations with a practically meaningful performance change threshold $\epsilon > 0$ and a small $\delta > 0$:}
\begin{equation}
\label{property2-insight3}
\mathrm{Pr}_{P_2 \sim \mathcal{D}_{\text{var}}(P_1)} \big\{ |\mathcal{H}(M, I, P_1) - \mathcal{H}(M, I, P_2)| > \epsilon \big\} \ge 1 - \delta.
\end{equation}
3) \emph{For a typical image $I$ and prompt $P$, the performance exhibits vulnerability to text noise $\eta_T$ that disrupts the intended semantics or image-text alignment relevant to the OOD task with a practically meaningful performance change threshold $w > 0$ and a small $\delta > 0$:}
\begin{equation}
\label{property3-insight3}
\mathrm{Pr}_{\eta_T} \big\{ \mathcal{H}(M, I, P) - \mathcal{H}(M, I, P + \eta_T) > w \big\} \ge 1 - \delta
\end{equation}
This insight formalizes crucial empirical observations from our study regarding the behavioral characteristics of VLM-based OOD detection methods. It reveals a striking asymmetry in robustness between the image and text modalities: while the method generally shows commendable resilience to common pixel-level noise and corruptions in the image input, its performance is highly sensitive to the specific wording of the text prompts and, more critically, appears notably vulnerable to various forms of detrimental text perturbations. This empirical finding, systematically documented in our work, underscores potential limitations in the robustness of the VLM's text encoding or its cross-modal alignment mechanism when faced with variations or targeted attacks in the textual domain. This suggests that careful prompt engineering, robust prompt selection strategies, and consideration of text-based adversarial threats are critical factors affecting the reliability of such methods in practice.

\subsection{Robustness to Image Perturbations}
\label{subsec:image robustness}
\vspace{-5pt}
We first analyze the robustness of VLM-based OOD detection when the input image is subjected to various forms of noise and common corruptions. Understanding this is important for assessing the reliability of the method in real-world scenarios where images may be degraded. Insight~\hyperref[insight3]{3} (1) suggests that performance should remain relatively stable under bounded image perturbations. Figure~\ref{fig:image_robustness_curves_appendix} visually supports this property. This observation supports the notion that VLM features capture high-level semantic content that is relatively invariant to low-level pixel noise or common visual degradations. Further details are provided in Appendix~\ref{appendix:subsubsec:image_robustness_details_appendix}.

\subsection{Sensitivity to Prompt Variations}
\label{subsec:prompt sensitivity}
\vspace{-5pt}
Unlike single-modal image methods, VLM-based OOD detection relies heavily on textual prompts to define concepts. The exact wording and structure of these prompts can vary significantly while ostensibly referring to the same underlying concept. Insight~\hyperref[insight3]{3} (2) points to a potential sensitivity of performance to such variations, and Figure~\ref{fig:prompt_sensitivity_scatter_appendix} illustrates the results stemming from this variation. This demonstrates that prompt engineering is not a trivial detail but a critical factor influencing the effectiveness of VLM-based OOD detection, highlighting the method's sensitivity to the textual input modality.

\subsection{Vulnerability to Text Perturbations}
\label{subsec:text vulnerability}
\vspace{-5pt}
Beyond simple variations, text prompts can also be subjected to malicious or accidental perturbations that are specifically designed to disrupt their intended meaning or the VLM's processing. Insight~\hyperref[insight3]{3} (3) suggests that VLM-based OOD detection may be particularly vulnerable to such textual manipulations, presenting a stark contrast to its relative image robustness. We discuss this in detail in Appendix~\ref{appendix:subsubsec:text_perturbations_details_appendix}. This suggests that the VLM's text encoding or the image-text alignment process might be more fragile in the face of targeted semantic or adversarial disruptions in the textual domain compared to the visual domain.

\vspace{-.05in}
\begin{center}
\fcolorbox{black}{green!10}{\parbox{0.98\linewidth}{\textit{\textbf{Takeaway: } VLM-based OOD detection's robustness is asymmetric: strong against image noise but highly fragile concerning textual prompts, a key vulnerability.}}}
\end{center}
\vspace{-.05in}


\section{Conclusion}
\label{sec:conclusion}

Our systematic empirical analysis validates how specific VLM embedding space properties, leveraged by ID and OOD prompts, enable effective out-of-distribution detection (Insight~\hyperref[insight1]{1}) and confer a decided advantage over unimodal methods in identifying semantic novelty (Insight~\hyperref[insight2]{2}). More critically, we uncover a significant robustness asymmetry (Insight~\hyperref[insight3]{3}): VLM-based OOD exhibits resilience to image noise but a pronounced sensitivity to prompt wording and a concerning vulnerability to textual perturbations. This identifies the textual modality as a key point of fragility, underscoring an urgent need for advancements in robust prompt engineering and text processing to ensure reliable VLM deployment. These empirically-grounded findings provide crucial guidance for developing more dependable and trustworthy VLM systems. Our study's scope and limitations are detailed in Appendix~\ref{appendix:sec:limitations_appendix}.

\newpage

\bibliographystyle{plainnat}
\bibliography{refs}

\newpage
\appendix

\section{Related Work}
\label{appendix:sec:related work}

Our work is situated at the intersection of Out-of-Distribution (OOD) detection and Vision-Language Models (VLMs). We briefly review relevant literature in these areas, specifically to highlight how our systematic empirical analysis of VLM-based OOD mechanisms, advantages, and sensitivities distinguishes itself and contributes to the existing body of knowledge.

\subsection{Single-modal Out-of-Distribution Detection}
\label{appendix:subsec:ood_detection_related}
Visual Out-of-distribution (OOD) detection \citep{survey1, survey2, yang2022openood}, a significant area in machine learning systems and computer vision, involves discerning images that deviate from the training data distribution, often due to lacking specific semantic information expected by the model. Following a prolonged period of development, this field has evolved extensive methodologies primarily focused on single-modality data. The foundational paradigm for OOD detection within neural networks was outlined by Hendrycks \textit{et al.} \citep{hendrycks2016baseline}, which spurred numerous subsequent developments. These can be broadly categorized into: output-based methods \citep{bendale2016towards,devries2018learning,du2022vos,hein2019relu,hsu2020generalized,huang2021mos,liu2020energy,sun2022dice}, which often leverage classifier confidence scores like Maximum Softmax Probability (MSP); density-based methods \citep{abati2019latent,zong2018deep,deecke2019image,sabokrou2018adversarially,pidhorskyi2018generative}, which aim to model the density of in-distribution data; distance-based methods \citep{lee2018simple,ming2022cider,ren2021simple,sun2022out}, which measure distances in feature space; reconstruction-based methods \cite{li2023rethinking,denouden2018improving,zhou2022rethinking,yang2022out}, which rely on autoencoders or similar models; and gradient-based methods \citep{liang2017enhancing,huang2021importance,igoe2022useful}, such as ODIN, which often use input pre-processing or temperature scaling.

Some works \citep{morteza2022provable, Fang2022IsOD, Bitterwolf2022BreakingDO} have also provided enhanced theoretical analyses for single-modal OOD detection. Recent studies continue to re-examine and address the OOD detection problem from various perspectives within the single-modal paradigm. For instance, Ammar \textit{et al.} \citep{ammar2024neco} proposed utilizing "neural collapse" and the geometric properties of principal component spaces to identify OOD data. Lu \textit{et al.} \citep{lu2023learning} modeled each class with multiple prototypes to learn more compact sample embeddings, thereby enhancing OOD detection capabilities. While these single-modal methods have made significant progress, particularly for OOD data characterized by noticeable low-level statistical shifts or residing outside the learned manifold of ID data, they often rely heavily on features learned solely from ID data for a specific task. As our work explores, this can limit their effectiveness in detecting more nuanced semantic novelty. Our analysis, therefore, focuses on the distinct paradigm offered by VLMs, which leverage rich, cross-modally learned semantic spaces.

\subsection{Vision-Language Models}
\label{appendix:subsec:vlms_related}
Large-scale pre-trained Vision-Language Models, most notably CLIP \cite{CLIP}, along with others like ALIGN \cite{ALIGN} and SigLIP \cite{zhai2023sigmoid}, have revolutionized cross-modal understanding and zero-shot learning. Trained on massive datasets of image-text pairs using a contrastive objective (or similar alignment strategies), VLMs learn a powerful joint embedding space where semantically related images and texts are mapped into close proximity. This shared space enables remarkable zero-shot transfer capabilities across a wide range of vision and language tasks without requiring task-specific fine-tuning \citep{yang2023diffusion,li2023blip,kirillov2023segment,radford2023robust,lin2023magic3d,fang2023eva}. The ability to connect visual content with high-level semantic concepts defined by text is the core strength of VLMs, setting them apart from models trained solely on single-modal data. Our work is built upon these powerful VLM architectures and specifically provides a systematic empirical investigation into how the inherent properties of their learned joint embedding space directly facilitate the task of OOD detection.

\subsection{VLM-based OOD Detection}
\label{appendix:subsec:vlm_ood_related}
The strong zero-shot capabilities and semantic understanding inherent in VLMs have naturally led to their increasing application in OOD detection. An early idea, proposed by Fort \textit{et al.} \citep{fort2021exploring}, involved utilizing large-scale pretrained multimodal models by relying on outlier class names without accompanying images. Many initial VLM-based OOD methods adapted the VLM's zero-shot classification mechanism: they computed the similarity between an image embedding and embeddings of textual prompts representing ID classes, using the maximum similarity or related scores as an OOD indicator \cite{zhou2022conditional}. MCM (Maximum Concept Matching) \citep{MCM} was a key work that, building on language-vision representations, utilized temperature scaling and the maximum predicted softmax value as the OOD score, establishing a foundational paradigm for CLIP-based OOD detection that moved away from traditional sample assumptions.

Building upon these principles, recent works have focused on enhancing VLM-based OOD detection, often by incorporating textual prompts beyond just ID class labels. This includes using prompts representing abstract, "out-of-distribution," or "negative" concepts alongside ID prompts. The core idea, which our paper empirically analyzes in depth (Insight~\hyperref[insight1]{1}, Property~\ref{property3-insight1}), is to utilize the relative affinities of an image to both ID and these auxiliary OOD/negative concepts within the VLM space to derive a more discriminative OOD score. For instance, GL-MCM \citep{glmcm} enhanced the MCM approach by aligning global and local visual-textual features from CLIP. Other works like ZOC \citep{ZOC} employed models to generate candidate unknown class names. LSN \citep{LSN}, NegPrompt \citep{NegPrompt}, CMA\citep{lee2025concept}, and NegLabel \citep{NegLabel} all explicitly introduced negative prompts, leveraging techniques such as prompt learning or sampling to improve OOD detection effectiveness. LoCoOp \citep{LoCoOp} explored few-shot OOD detection within a prompt learning framework.

The specific framework empirically analyzed in this paper—which computes a unified score based on the normalized similarity to ID prompts relative to the combined set of ID and OOD prompts (conceptually similar to approaches in \cite{lee2025concept, NegLabel} that also leverage relative affinities)—is a prominent example of this advanced, post-hoc paradigm that typically obviates the need for additional learning or external data beyond the pre-trained VLM and prompt definitions. Despite the empirical successes of such methods, a systematic analytical understanding of these characteristics of vision language models and robustness features on out-of-distribution detection tasks remains underdeveloped. Our work provides this necessary comprehensive empirical dissection.

\subsection{Analysis and Robustness of VLMs}
\label{appendix:subsec:analysis_vlms_related}
There is a growing and critical body of research dedicated to understanding the properties and behaviors of large pre-trained models like VLMs. Works have analyzed the structure of CLIP's embedding space \cite{goel2022cyclip, zhou2022conditional}, investigated its zero-shot transfer capabilities \cite{CLIP,zhou2022conditional}, and explored its biases \cite{goh2021multimodal, agarwal2021evaluating}. Additionally, the robustness of VLMs to various input perturbations—including image corruptions \cite{hendrycks2019benchmarking, engstrom2019exploring}, adversarial attacks on images \cite{zhao2023evaluating}, and variations or attacks on textual prompts \cite{wang2020cross, yang2023fine}—has been studied, primarily in the context of zero-shot classification and other downstream tasks. However, the specific implications of these VLM properties and robustness characteristics for the task of OOD detection, particularly the asymmetric sensitivity to image versus textual input variations within the OOD detection context, has not been systematically analyzed. Our study contributes to this area by providing an in-depth empirical investigation into the robustness and sensitivity of VLM-based OOD detection specifically.

\section{Experiment Details}
\label{appendix:sec:experiment Details}

This appendix section provides comprehensive details regarding the experimental setup used to obtain the results presented in the main paper (Sections~\ref{sec:mechanisms}, \ref{sec:advantage-single-modal}, \ref{sec:analysis}) and other appendix sections (Appendix~\ref{appendix:sec:additional results}). These details are essential for ensuring the reproducibility of our empirical findings.

\subsection{Software and Hardware}
\label{appendix:subsec:software and hardware}

All experimental implementations were conducted using PyTorch version 2.0.1 and Python version 3.8. Model training (if applicable, e.g., for baselines or any fine-tuning) and all OOD detection evaluations were performed on a server equipped with 4 NVIDIA RTX 4090 GPUs and an Intel Xeon(R) Platinum 8352V CPU.

\subsection{Metrics}
\label{appendix:subsec:metrics}
For evaluation, we use the following metrics: (1) the false positive rate (FPR95) of OOD samples when the true positive rate of in-distribution samples is at 95\%, (2) the area under the receiver operating characteristic curve (AUROC). All evaluation outcomes for our method are derived from the average of three experiments.

\subsection{Datasets}
\label{appendix:subsec:dataset}

We primarily use ImageNet-1k \cite{imagenet1k} as our in-distribution (ID) dataset. ImageNet-1k consists of $1.28$ million training images and $50$ thousand validation images across 1000 classes. For OOD detection evaluation, we use the standard ImageNet-1k validation set as the ID test set.

For out-of-distribution (OOD) evaluation in our main performance comparison (Section~\ref{sec:advantage-single-modal}, Appendix~\ref{appendix:subsec:detailed performance evaluation}) and robustness/sensitivity analyses (Section~\ref{sec:analysis}, Appendix~\ref{appendix:subsec:robustness_sensitivity_details_appendix}), we use four common OOD benchmark datasets for ImageNet:
\begin{itemize}
\item \textbf{iNaturalist} \cite{van2018inaturalist}: A dataset of fine-grained categories (e.g., specific species of animals, plants, fungi). We use approximately 10 thousand test images from classes disjoint from ImageNet-1k. It represents semantic OOD.
\item \textbf{SUN} \cite{xiao2010sun}: A dataset of scene categories. We use approximately 10 thousand test images from scene classes disjoint from ImageNet-1k. It represents semantic OOD (different high-level concepts from objects).
\item \textbf{Places} \cite{zhou2017places}: Another large-scale scene dataset. We use approximately 10 thousand test images from scene classes disjoint from ImageNet-1k. Similar to SUN, it represents semantic OOD.
\item \textbf{Textures} \cite{textures}: A dataset consisting of textural patterns. We use approximately 5.6 thousand test images. It primarily represents non-semantic or textural OOD.
\end{itemize}

For the hierarchical semantic alignment analysis (Appendix~\ref{appendix:subsubsec:hierarchical_semantic_alignment_appendix}), we use four specialized, finer-grained classification datasets as ID datasets:
\begin{itemize}
\item \textbf{Oxford-IIIT Pets} \cite{pet}: 37 categories of pets.
\item \textbf{Food-101} \cite{food101}: 101 categories of food items.
\item \textbf{CUB-200} \cite{cub200}: 200 categories of birds.
\item \textbf{Stanford Dogs} \cite{dogs}: 120 categories of dog breeds.
\end{itemize}

When using these specialized datasets as ID, OOD samples were drawn from a combination of the other specialized datasets or standard OOD benchmarks depending on the specific experiment setup (details provided in Appendix \ref{appendix:subsec:dataset}). All datasets were preprocessed following standard practices, including resizing images to $224 \times 224$ pixels and normalizing pixel values.

\subsection{Models and Implementation}
\label{appendix:subsec:model and implementation}

This subsection details the models employed for both the VLM-based OOD method central to our analysis and the single-modal baselines used for comparison, along with implementation specifics for these baselines.

\subsubsection{Models}
\label{appendix:subsubsec:models}

For our VLM-based OOD detection method, we primarily utilize various pre-trained Vision-Language Models available through public model hubs. The main analyses presented in Sections~\ref{sec:mechanisms}, \ref{sec:advantage-single-modal}, and \ref{sec:analysis} predominantly focus on \textbf{CLIP} \cite{CLIP} variants, specifically \texttt{RN50} (ResNet-50 based) and \texttt{ViT-B/16} (Vision Transformer B/16 based). These are standard models released by OpenAI and pre-trained on the WebImageText dataset. These models are utilized strictly off-the-shelf in a zero-shot manner for OOD detection, meaning no task-specific fine-tuning is performed on ImageNet-1k (when used as ID) or any of the OOD datasets. This ensures a fair evaluation of their intrinsic zero-shot OOD capabilities.

In Appendix~\ref{appendix:subsubsec:vlm_backbone_appendix} (Choice of VLM Backbone), we extend our analysis to evaluate other VLM backbones to understand the generality of our findings. These include larger CLIP models like \texttt{CLIP-ViT-L/14}, and models from the OpenCLIP \cite{openclip} suite such as \texttt{OpenCLIP-ViT-B/32} and \texttt{OpenCLIP-ViT-B/16} (pre-trained on variants of the LAION dataset \cite{schuhmann2021laion}), as well as ALIGN-Base \cite{ALIGN} (pre-trained by Google). These models are used as interchangeable image and text encoders within our VLM-based OOD framework defined in Appendix~\ref{appendix:subsec:algorithm}.

For the single-modal baseline methods, we use standard image classification models pre-trained on ImageNet-1k: a ResNet50 \cite{he2016deep} and a Vision Transformer (ViT-B/16) \cite{dosovitskiy2020image}. These models serve as feature extractors and provide the logits or features used by the various single-modal OOD scoring functions detailed below. These models are also used off-the-shelf without any further fine-tuning specifically for OOD detection, ensuring a fair comparison in what can be considered a zero-shot or transfer setting for OOD detection from an ImageNet-1k pre-trained model.

\subsubsection{Implementation of Other Methods}
\label{appendix:subsubsec:other methods}

For comparison with the VLM-based approach, we implement several widely recognized single-modal image OOD detection methods based on features or logits from the pre-trained ImageNet-1k classifiers (ResNet50 and ViT-B/16). These methods typically operate on features extracted from a standard image classification model trained on the in-distribution (ID) data. The OOD score $S(x)$ for an input image $x$ is computed based on the model's output (logits or features). For consistency, we define the score $S(x)$ such that higher values indicate an ID sample, and lower values indicate an OOD sample. The specific scoring functions for MSP\cite{hendrycks2016baseline}, MaxLogit\cite{hendrycks2022scaling}, ODIN\cite{liang2017enhancing}, and Energy-based Methods\cite{liu2020energy} are detailed below:

\paragraph{Maximum Softmax Probability (MSP)}
The MSP method, a simple yet effective baseline, utilizes the confidence of a standard classifier trained on ID data. It posits that ID samples will have high confidence (maximum softmax probability) on their true class, while OOD samples will result in lower confidence across all classes. The OOD score is simply the maximum probability output by the softmax layer:
\begin{equation}
S_{\text{MSP}}(x) = \max_{y \in C_{\text{ID}}} P(y|x),
\end{equation}
where $P(y|x) = \frac{\exp(z_y)}{\sum_{y' \in C_{\text{ID}}} \exp(z_{y'})}$ is the softmax probability for class $y$, and $z_y$ are the logits from the classifier's output layer.

\paragraph{MaxLogit}
Similar to MSP, the Max Logit method directly uses the raw logits output by the classifier. The assumption is that for ID samples, the maximum logit value will be higher compared to OOD samples, which do not strongly activate any of the known class outputs. The OOD score is the maximum value among the logits:
\begin{equation}
S_{\text{MaxLogit}}(x) = \max_{y \in C_{\text{ID}}} z_y,
\end{equation}
where $z_y$ are the logits for class $y$.

\paragraph{ODIN}
ODIN enhances the MSP score by applying temperature scaling to the logits and introducing a small, gradient-based adversarial perturbation to the input image. This perturbation is designed to increase the confidence of the predicted class for an input, which is hypothesized to push OOD samples further away from ID decision boundaries in the logit space. The score is the maximum softmax probability after applying temperature scaling and input perturbation:
\begin{equation} 
S_{\text{ODIN}}(x) = \max_{y \in C_{\text{ID}}} \frac{\exp(\hat{z}_y/\tau)}{\sum_{y' \in C_{\text{ID}}} \exp(\hat{z}_{y'}/\tau)},
\end{equation}
where $\hat{z} = z(x')$ are the logits for the perturbed input $x' = x - \epsilon \cdot \text{sign}(\nabla_x \log \frac{\exp(z_{\hat{y}}/\tau)}{\sum_{y'} \exp(z_{y'}/\tau)})$, $\hat{y} = \arg\max_{y \in C_{\text{ID}}} z_y$ is the predicted class based on original logits, $\tau$ is the temperature parameter, and $\epsilon$ is the perturbation magnitude.

\paragraph{Energy-based Methods}
Energy-based OOD detection methods define an energy function based on the output logits. Lower energy values are associated with ID data, while higher energy values indicate OOD data. The energy score is defined as the negative LogSumExp of the logits. To align with our convention (higher score = ID), we use the negative of the energy score:
\begin{equation}
S_{\text{Energy}}(x) = \log \sum_{y \in C_{\text{ID}}} \exp(z_y),
\end{equation}
where $z_y$ are the logits for class $y$.

\subsection{VLM-based OOD Detection Algorithm Details}
\label{appendix:subsec:algorithm}

This subsection provides the precise algorithmic steps for the VLM-based OOD detection method analyzed in this paper, which utilizes both ID and OOD prompts. The core pipeline, previously introduced conceptually in Section~\ref{sec:pre} and illustrated in Figure~\ref{fig: framework}, involves encoding images and prompts into the VLM's shared embedding space and computing a unified score based on their similarities. The specific process is detailed below and summarized in Algorithm~\ref{alg:Unified_OOD_Detection_Pipeline}.

We investigate the scenario where both in-distribution (ID) and out-of-distribution (OOD) prompts (sometimes also referred to as neutral or agent prompts) are used in conjunction with CLIP’s feature space. Figure~\ref{fig: framework} shows a unified framework, where the \textbf{image encoder} $g(\cdot)$ extracts a visual feature $\mathbf{v} \in \mathbb{R}^d$ and the \textbf{text encoder} $f(\cdot)$ extracts features for a set of textual prompts, denoted $\{\mathbf{c}_1, \dots, \mathbf{c}_K\}$ for the $K$ ID categories, and $\{\mathbf{c}_{K+1}, \dots, \mathbf{c}_{K+M}\}$ for $M$ additional OOD (or neutral) concepts.

\paragraph{Constructing ID and OOD Prompts.}
Let $\mathcal{Y}_{\text{in}}$ denote the set of $K$ Original ID labels. For each $y_i \in \mathcal{Y}_{\text{in}}$, we create a textual prompt $t_i = \text{Converter}_{id}{(y_i)}$ (e.g., \texttt{a photo of a \{label\}}), which is encoded by the text encoder $f(\cdot)$ to generate ID textual features:
\[
\{\mathbf{c}_1, \dots, \mathbf{c}_K\}, \quad \mathbf{c}_k = f(t_k).
\]
In addition, we construct $M$ OOD prompts $T_{ood} = \text{Converter}_{ood}{(\mathcal{Y}_{\text{in}})}$ to represent abstract or unrelated concepts, such as ``\texttt{an unrelated object}'' or ``\texttt{a random scene}''. These prompts generate OOD textual features:
\[
\{\mathbf{c}_{K+1}, \dots, \mathbf{c}_{K+M}\}, \quad \mathbf{c}_{k+n} = f(t_{k+n}).
\]
These prompts aim to represent abstract or unrelated concepts (e.g., ``\texttt{an unrelated object}'', ``\texttt{a random scene with no known objects}'', etc.). The combined text features from both ID and OOD prompts can be leveraged to better separate ID from OOD inputs.

\paragraph{Unified Scoring with ID and OOD Prompts.}
Let $\mathbf{v} = g(\mathbf{x})$ be the visual feature of an image input $x$. We compute the cosine similarity
\begin{equation}
\label{eq:sim}
s_i(x) = \mathrm{sim}(\mathbf{v}, \mathbf{c}_i) = \frac{\mathbf{v} \cdot \mathbf{c}_i}{\|\mathbf{v}\| \|\mathbf{c}_i\|}, \quad i = 1, \dots, K+M.
\end{equation}
Define $\hat{k} = \arg\max_{1 \leq i \leq K} s_i(x)$ as the ID concept best matching $\mathbf{v}$. We then adapt the Maximum Concept Matching (MCM)\cite{MCM} strategy: 
\begin{equation}
\label{eq:score_id}
S_{\mathrm{ID}}(x) 
\;=\;
\frac{\exp\bigl(s_{\hat{k}}(x)/\tau\bigr)}
{\sum_{i=1}^{K} \exp\bigl(s_i(x)/\tau\bigr)},
\end{equation}
by including OOD prompts in the normalization:
\begin{equation}
\label{eq:score_id_ood}
S_{\mathrm{ID+OOD}}(x) 
\;=\;
\frac{\exp\bigl(s_{\hat{k}}(x)/\tau\bigr)}
{\sum_{i=1}^{K+M} \exp\bigl(s_i(x)/\tau\bigr)},
\end{equation}
where $\tau$ is a temperature hyperparameter. Note that the \emph{numerator} only uses the ID prototype $\mathbf{c}_{\hat{k}}$, whereas the \emph{denominator} incorporates all $K+M$ prototypes (ID + OOD). We then classify:
\begin{equation}
\label{eq:final_decision}
G_{\mathrm{ID+OOD}}(\mathbf{x})
\;=\;
\begin{cases}
\text{ID}, & S_{\mathrm{ID+OOD}}(x) \;\ge\; \lambda,\\
\text{OOD}, & \text{otherwise},
\end{cases}
\end{equation}
where $\lambda$ is the threshold, and examples below $\lambda$ are considered OOD inputs. Hence, $S_{\mathrm{ID+OOD}}(x)$ acts as a ``unified score'' that leverages both ID and OOD prompts in a single calculation. When $x$ is OOD, it will often align more with the OOD prompts in the denominator, thus lowering the overall fraction in~\eqref{eq:score_id_ood}. 

\begin{algorithm}[h]
\caption{Unified VLM-based OOD Detection Pipeline}
\label{alg:Unified_OOD_Detection_Pipeline}
\begin{algorithmic}[1]
\State \textbf{Require:} ID labels $\{ \text{cat}, \text{dog}, \ldots, \text{car} \}$, ID images $\{ \textbf{x}_{id} \}$, OOD images $\{ \textbf{x}_{ood} \}$
\State \textbf{Ensure:} OOD detection scores for input images

\Statex
\State \textbf{Initialize:} 
    $g(\cdot)$: image encoder, $f(\cdot)$: text encoder, Define ID and OOD prompts $T_{\text{id}}, T_{\text{ood}}$ via converter from original ID labels

\Statex
\State \textbf{Process Inputs:}
\For{each image $x \in \{ \textbf{x}_{id}, \textbf{x}_{ood} \}$}
    \State $v \gets g(x)$ \Comment{Visual representation}
\EndFor

\Statex
\State \textbf{Encode Prompts:}
\For{each prompt $t \in \{ T_{\text{id}}, T_{\text{ood}} \}$}
    \State $c \gets f(t)$ \Comment{Text embeddings}
\EndFor

\Statex
\State \textbf{Compute Similarity:}
\For{each image $x$}
    \State $s(x) \gets \left\{ \frac{v \cdot c_i}{\|v\| \|c_i\|} \right\}_{i=1}^{K+M}$ \Comment{Similarity scores}
    \State $\hat{k} \gets \arg\max_{1 \leq i \leq K} s_i(x)$
    \State $S_{\text{ID+OOD}}(x) \gets \frac{\exp\left( s_{\hat{k}}(x)/\tau \right)}{\sum_{i=1}^{K+M} \exp\left( s_i(x)/\tau \right)}$
\EndFor

\Statex
\State \textbf{Output:} OOD detection scores
\end{algorithmic}
\end{algorithm}

\subsection{Hyperparameters}
\label{appendix:subsec:hyperparameters}

For the VLM-based OOD detection method (Algorithm~\ref{alg:Unified_OOD_Detection_Pipeline}), the primary tunable hyperparameter is the temperature $\tau$ in the scoring function (Equation~\ref{eq:score_id_ood}). Based on validation experiments exploring the sensitivity of performance to $\tau$ (detailed in Appendix~\ref{appendix:subsubsec:temperature_sensitivity_appendix}), we consistently found that a value of $\tau = 1.0$ yields strong and stable performance across various VLM backbones and OOD datasets. Therefore, this value is used for the main experiments presented in Section~\ref{sec:advantage-single-modal}, Section~\ref{sec:analysis}, and Appendices~\ref{appendix:subsec:detailed performance evaluation}, \ref{appendix:subsec:analysis of VLM}, and \ref{appendix:subsec:robustness_sensitivity_details_appendix}. 
The basic ID prompt template and the set of OOD prompts used in our main experiments followed the settings in \citet{NegLabel}, \citet{MCM} \citet{lee2025concept}.
Regarding the number of OOD prompts $M$, we follow the conclusion provided in \citet{lee2025concept} and set it to be equal to the number of ID prompts, i.e., $M=K$.
The threshold $\lambda$ for the final ID/OOD classification (Equation~\ref{eq:final_decision}) is not a fixed hyperparameter but is determined by the desired operating point on the ROC curve during evaluation.

For the single-modal ODIN baseline \cite{liang2017enhancing} (detailed in Appendix~\ref{appendix:subsubsec:other methods}), the key hyperparameters are its specific temperature $T_{odin}$ and the perturbation magnitude $\epsilon$. Following common practice and recommendations from the original paper, we set $T_{odin} = 1000$ and $\epsilon = 0.0014$ (adjusted for pixel values in $[0,1]$ range if necessary, typically $1.4/255$). These values are widely used and were found to provide competitive performance for ODIN on ImageNet-scale benchmarks. For other single-modal baselines such as MSP, MaxLogit, and the Energy-based score (with $T_{energy}=1.0$), there are typically no other hyperparameters to tune beyond the choice of the base pre-trained model.

\section{Additional Experimental Results}
\label{appendix:sec:additional results}
This Appendix section provides comprehensive details and additional experimental results supporting the key findings and insights presented in Sections~\ref{sec:mechanisms}, \ref{sec:advantage-single-modal}, and \ref{sec:analysis} of the main paper. We include further analysis, detailed quantitative results, and visualization specifics necessary for full understanding and potential reproducibility of our empirical observations.

\subsection{Detailed Performance Evaluation}
\label{appendix:subsec:detailed performance evaluation}
This subsection provides comprehensive quantitative results comparing the performance of the proposed VLM-based OOD detection method, which leverages both ID and OOD textual prompts, against representative single-modal image-based baselines. These results expand upon and provide the full data supporting the discussion and Insight~\hyperref[insight2]{2} presented in Section~\ref{sec:advantage-single-modal}. We evaluate methods using ImageNet-1k as the in-distribution dataset and four diverse out-of-distribution datasets.

\subsubsection{Full Comparison Tables}
\label{appendix:subsubsec:full comparison tables}
Table~\ref{table:appendix_full_performance_comparison} presents the complete performance results (FPR95 and AUROC) for all evaluated methods across the four OOD datasets, including the average performance. The table includes several widely-used single-modal OOD detection methods (detailed in Appendix~\ref{appendix:subsubsec:other methods}) applied to features from standard image classification models pre-trained on ImageNet-1k (ResNet50 and ViT-B/16), as well as our VLM-based method utilizing CLIP variants (CLIP-RN50 and CLIP-ViT-B/16) with the ID and OOD prompt scoring mechanism from Appendix~\ref{appendix:subsec:algorithm}.

As is evident from Table~\ref{table:appendix_full_performance_comparison}, the VLM-based methods consistently and significantly outperform the single-modal baselines across all evaluated OOD datasets and metrics. For instance, on average, CLIP-ViT achieves the best performance with an AUROC of 92.62\% and an FPR95 of 32.87\%. This is substantially better than the best-performing single-modal method on average (which, in this comprehensive table, is RN50-MaxLogit with an AUROC of 87.81\% and FPR95 of 52.60\%). This clear and consistent quantitative advantage strongly supports Insight~\hyperref[insight2]{2}, robustly demonstrating the superior capability of VLM-based methods for OOD detection in these challenging, semantically diverse benchmark scenarios. The VLM's ability to leverage textual prompts for semantic disambiguation appears to be a key factor in this enhanced performance.

\begin{table*}[h!] 
\centering
\resizebox{\textwidth}{!}{
    \begin{tabular}{@{}l *{10}{c} @{}}
        \toprule
        & \multicolumn{2}{c}{iNaturalist} & \multicolumn{2}{c}{SUN} & \multicolumn{2}{c}{Places} & \multicolumn{2}{c}{Textures} & \multicolumn{2}{c}{\textbf{Average}} \\
        \cmidrule(lr){2-3} \cmidrule(lr){4-5} \cmidrule(lr){6-7} \cmidrule(lr){8-9} \cmidrule(lr){10-11}
        \textbf{Method} & \scriptsize FPR95$\downarrow$ & \scriptsize AUROC$\uparrow$ & \scriptsize FPR95$\downarrow$ & \scriptsize AUROC$\uparrow$ & \scriptsize FPR95$\downarrow$ & \scriptsize AUROC$\uparrow$ & \scriptsize FPR95$\downarrow$ & \scriptsize AUROC$\uparrow$ & \scriptsize FPR95$\downarrow$ & \scriptsize AUROC$\uparrow$  \\
        \midrule
        \multicolumn{11}{l}{\textit{Single-modal Methods (ImageNet-1k Pre-trained Backbone)}} \\ 
        \textbf{ResNet50 Backbone} & & & & & & & & & & \\
        MSP \cite{hendrycks2016baseline}   & 52.20 & 87.77 & 71.40 & 79.67 & 71.50 & 80.37 & 66.50 & 78.59 & 65.40 & 83.10 \\
        MaxLogit \cite{hendrycks2022scaling} & 44.30 & 91.94 & 54.20 & 87.36 & 63.00 & 84.09 & 48.90 & 87.56 & 52.60 & 87.81 \\
        Energy \cite{liu2020energy} & 50.30 & 90.68 & 56.60 & 86.83 & 62.90 & 84.24 & 49.60 & 86.65 & 53.35 & 86.65 \\
        ODIN \cite{liang2017enhancing}   & 42.10 & 91.83 & 57.00 & 86.44 & 64.40 & 83.26 & 47.00 & 87.43 & 52.63 & 87.59 \\
        \addlinespace 
        \textbf{ViT-B/16 Backbone} & & & & & & & & & & \\
        MSP \cite{hendrycks2016baseline}    & 49.60 & 89.56 & 63.80 & 83.72 & 67.10 & 82.51 & 66.00 & 81.00 & 61.63 & 83.34 \\
        MaxLogit \cite{hendrycks2022scaling}& 46.50 & 88.48 & 60.50 & 83.03 & 65.20 & 79.20 & 62.60 & 80.14 & 58.70 & 81.44 \\
        Energy \cite{liu2020energy} & 47.70 & 87.21 & 54.70 & 82.61 & 61.50 & 80.47 & 54.40 & 80.72 & 53.33 & 82.51 \\
        ODIN \cite{liang2017enhancing}   & 45.80 & 88.25 & 55.20 & 83.93 & 65.50 & 78.79 & 59.10 & 79.66 & 58.65 & 82.65 \\
        \addlinespace 
        \midrule
        \multicolumn{11}{l}{\textit{Vision-Language Model Methods (Zero-shot)}} \\
        CLIP-RN50 & 24.05 & 95.26 & 37.54 & 91.32 & 44.51 & 88.92 & 39.24 & 90.85 & 36.33 & 91.58  \\
        CLIP-ViT-B/16   & 23.80 & 95.46 & 30.52 & 93.40 & 29.82 & 93.18 & 47.38 & 88.48 & 32.87 & 92.62 \\
        \bottomrule
    \end{tabular}
} 
\caption{Comprehensive performance comparison (FPR95$\downarrow$ and AUROC$\uparrow$) of VLM-based OOD detection methods against various single-modal baselines using different backbone architectures, on four OOD datasets with ImageNet-1k as in-distribution. VLM-based methods demonstrate consistently superior results.}
\label{table:appendix_full_performance_comparison}
\end{table*}

\subsubsection{Additional Visualizations}
\label{appendix:subsubsec:additional visualizations}

\begin{figure*}[h!]
\begin{center}
\includegraphics[width=0.7\textwidth]{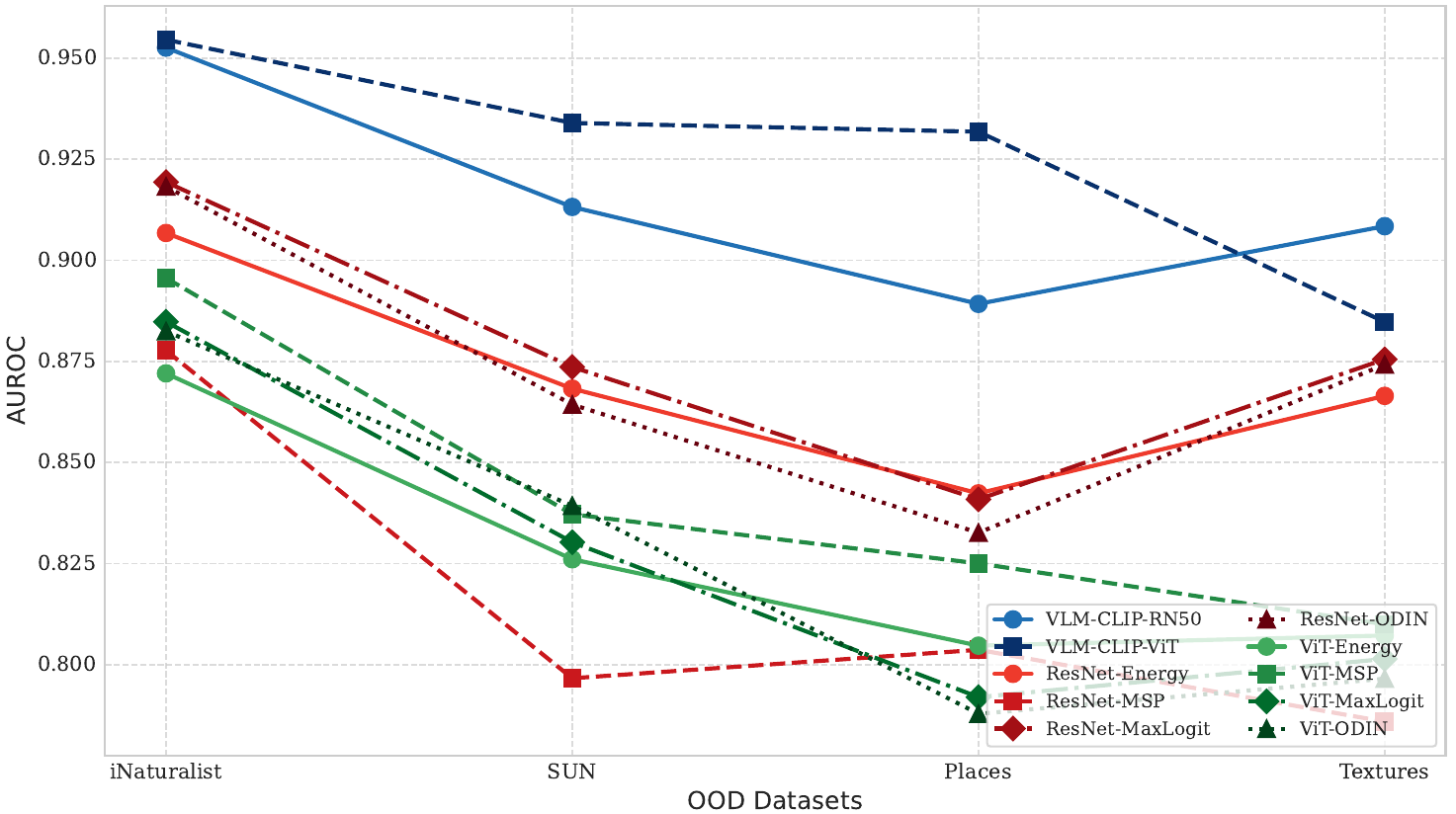} 
\end{center}
\caption{AUROC performance across different OOD Datasets for all evaluated VLM-based and Single-modal methods (from Table~\ref{table:appendix_full_performance_comparison}). Higher is better. VLM-based methods consistently demonstrate superior AUROC.}
\label{fig:appendix_aurocs_by_dataset_detailed}
\vspace{-.05in}
\end{figure*}

\begin{figure*}[h!]
\begin{center}
\includegraphics[width=0.7\textwidth]{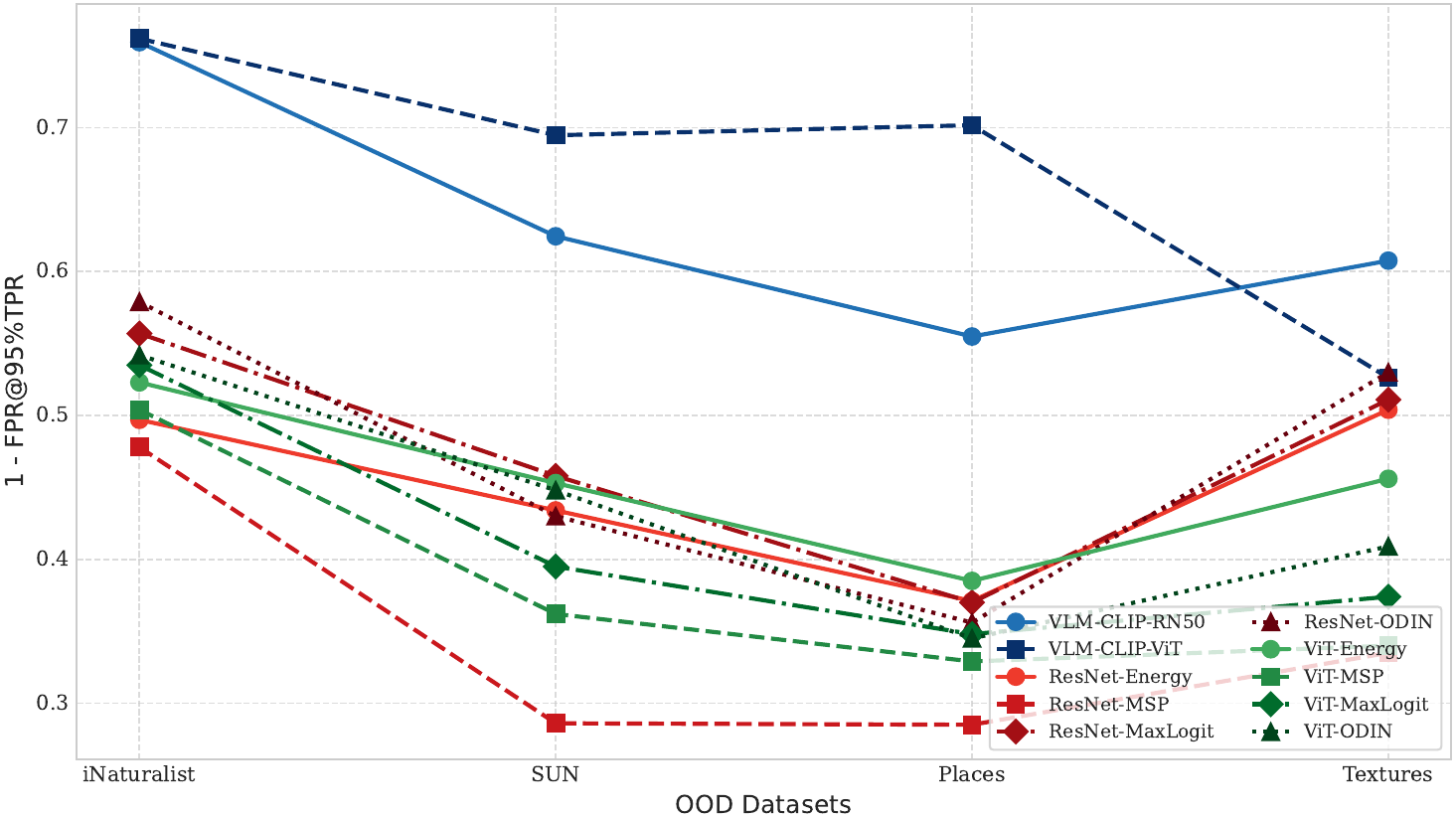} 
\end{center}
\caption{1-FPR95 performance across different OOD Datasets for all evaluated VLM-based and Single-modal methods (from Table~\ref{table:appendix_full_performance_comparison}). Higher is better. VLM-based methods consistently show lower FPR95.}
\label{fig:appendix_fpr95s_by_dataset_detailed}
\vspace{-.05in}
\end{figure*}

To provide a clearer visual comparison that complements Table~\ref{table:appendix_full_performance_comparison}, Figure~\ref{fig:appendix_aurocs_by_dataset_detailed} and Figure~\ref{fig:appendix_fpr95s_by_dataset_detailed} plot the AUROC and FPR95 performance respectively across the four OOD datasets for all methods listed in the comprehensive table. These line plots vividly illustrate the consistent and substantial lead of the two VLM-based methods (CLIP-RN50 and CLIP-ViT-B/16) over the array of single-modal baselines across different types of OOD data. The VLM methods not only achieve higher absolute performance but also exhibit more stable behavior across datasets relative to single-modal methods, some of which show significant performance drops on certain OOD datasets (e.g., several ResNet50-based methods on SUN and Places).

\begin{figure*}[h!]
\begin{center}
\includegraphics[width=0.7\textwidth]{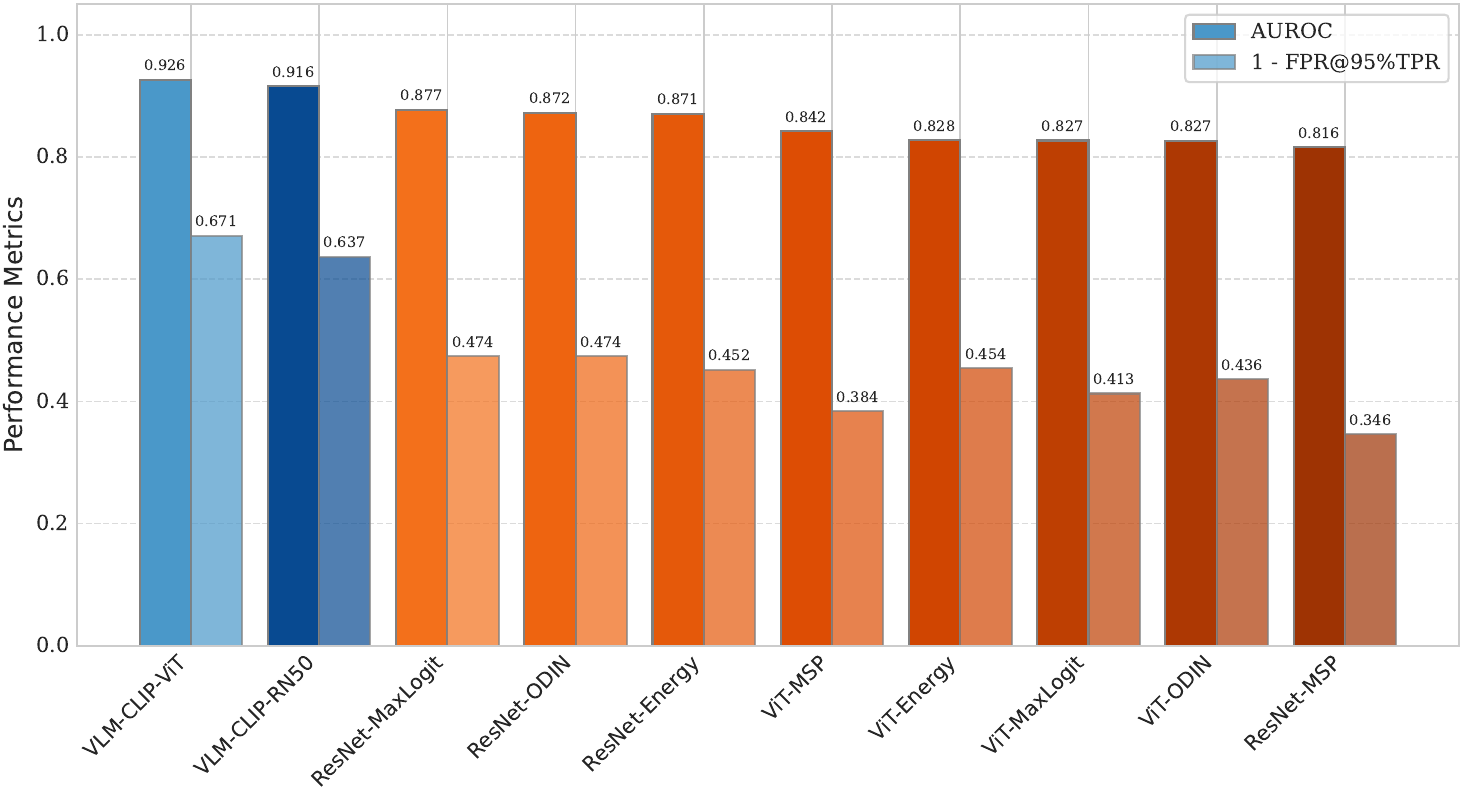} 
\end{center}
\caption{Average performance comparison (AUROC and 1-FPR95, so higher is better for both) of VLM-based vs. Single-modal OOD Detection methods across all OOD datasets, summarizing Table~\ref{table:appendix_full_performance_comparison}. VLM-based methods achieve significantly higher average performance.}
\label{fig:appendix_avg_performance_bar_detailed}
\vspace{-.05in}
\end{figure*}

Figure~\ref{fig:appendix_avg_performance_bar_detailed} provides a bar chart visualizing the average AUROC and average (1-FPR95) across all four OOD datasets for all methods. This aggregated view clearly highlights the substantial overall performance gap between the VLM-based methods and the single-modal baselines, quantitatively summarizing the comprehensive results from Table~\ref{table:appendix_full_performance_comparison}.

\begin{figure*}[h!]
\begin{center}
\includegraphics[width=\textwidth]{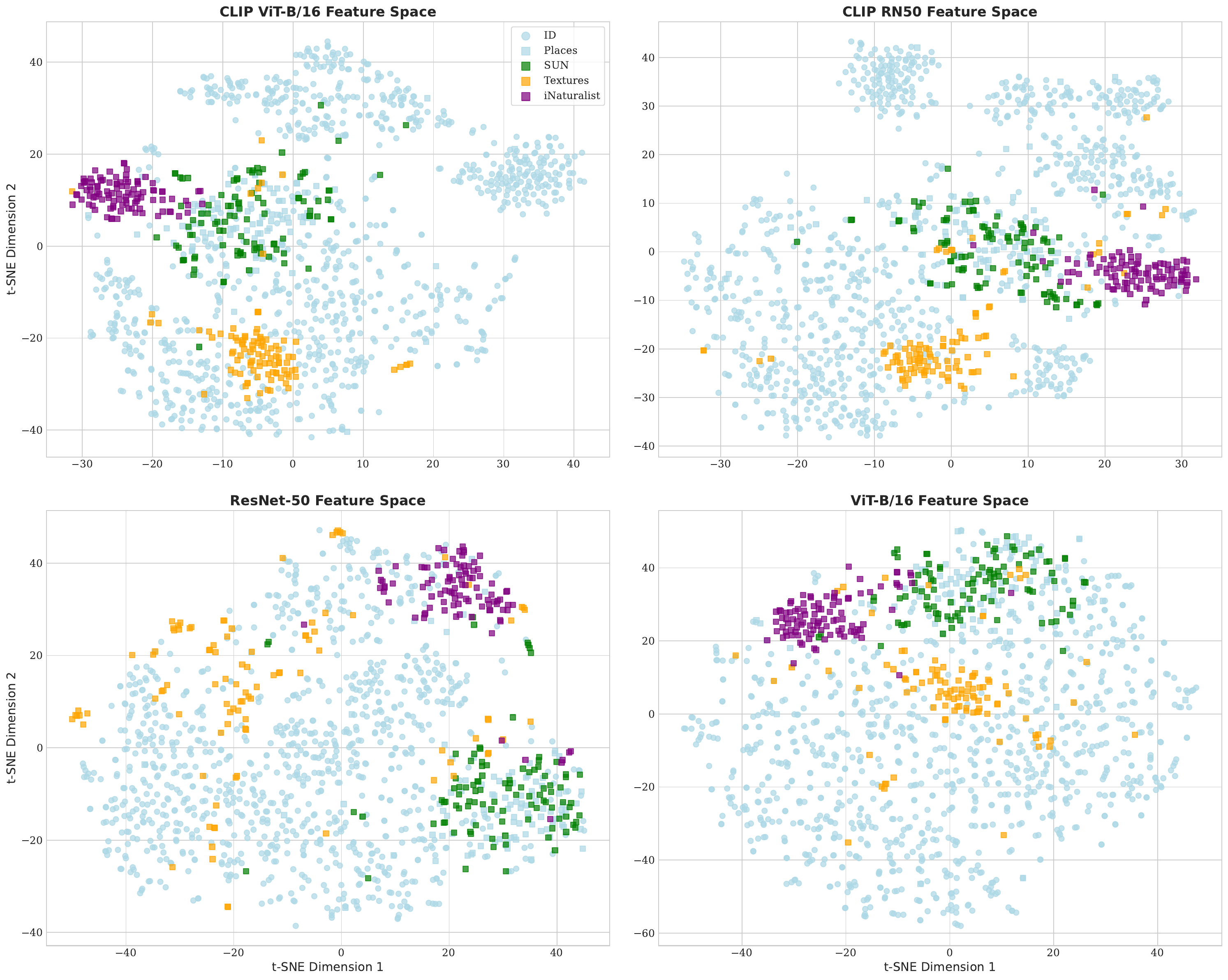} 
\end{center}
\caption{t-SNE visualizations of image embeddings for ID (ImageNet-1k, light blue) and OOD datasets (iNaturalist - purple, SUN - orange, Places - green, Textures - red) in the feature spaces of VLM models (CLIP-ViT-B/16, CLIP-RN50) and single-modal models (ResNet50, ViT-B/16). The VLM spaces qualitatively show more distinct clustering and separation for OOD data.}
\label{fig:appendix_model_comparison_tsne_detailed}
\vspace{-.05in}
\end{figure*}

Furthermore, Figure~\ref{fig:appendix_model_comparison_tsne_detailed} presents t-SNE visualizations of the image embeddings (from the respective model's feature extraction layer) for ID samples (ImageNet-1k) and samples from the four OOD datasets. We compare embeddings from representative VLM models (CLIP-ViT-B/16, CLIP-RN50) and single-modal models (ResNet50, ViT-B/16 trained on ImageNet). Qualitatively, these visualizations suggest that the VLM feature spaces (even before explicit prompt guidance) tend to exhibit a more semantically structured organization, where OOD datasets often form more separable clusters from the ID data, and also from each other, compared to the feature spaces of single-modal classifiers. This visual evidence, while qualitative, complements the quantitative results and supports the hypothesis that the VLM's contrastive pre-training fosters a semantic embedding space that is inherently better structured for distinguishing conceptual novelty—a key factor underpinning their superior OOD detection performance, especially against semantic shifts.

\subsection{Analysis of VLM Embedding Space Properties for OOD Detection}
\label{appendix:subsec:analysis of VLM}
This subsection provides detailed empirical evidence and visualizations supporting the claimed operational properties of the VLM joint image-text embedding space that enable effective zero-shot OOD detection, as formalized in Insight~\hyperref[insight1]{1} (Section~\ref{sec:mechanisms}). We analyze the alignment of image embeddings with text prompts and the relative positioning and separability of in-distribution (ID) and out-of-distribution (OOD) data when guided by these prompts. These empirical characterizations form the basis for understanding \textit{how} VLM-based OOD detection functions.

\subsubsection{ID Classification Alignment Details}
\label{appendix:subsubsec:id_classification_alignment_appendix} 

The first fundamental property we systematically validate is the ID Classification Alignment: for an ID image, its embedding is expected to have maximum similarity to the text embedding of its true class prompt compared to other ID class prompts. This property, a direct result of the VLM's contrastive training, underpins the model's capability to associate images with their correct semantic concepts in the shared embedding space, which is a prerequisite for distinguishing them from OOD concepts.

\begin{figure*}[h!]
\begin{center}
\includegraphics[width=0.7\textwidth]{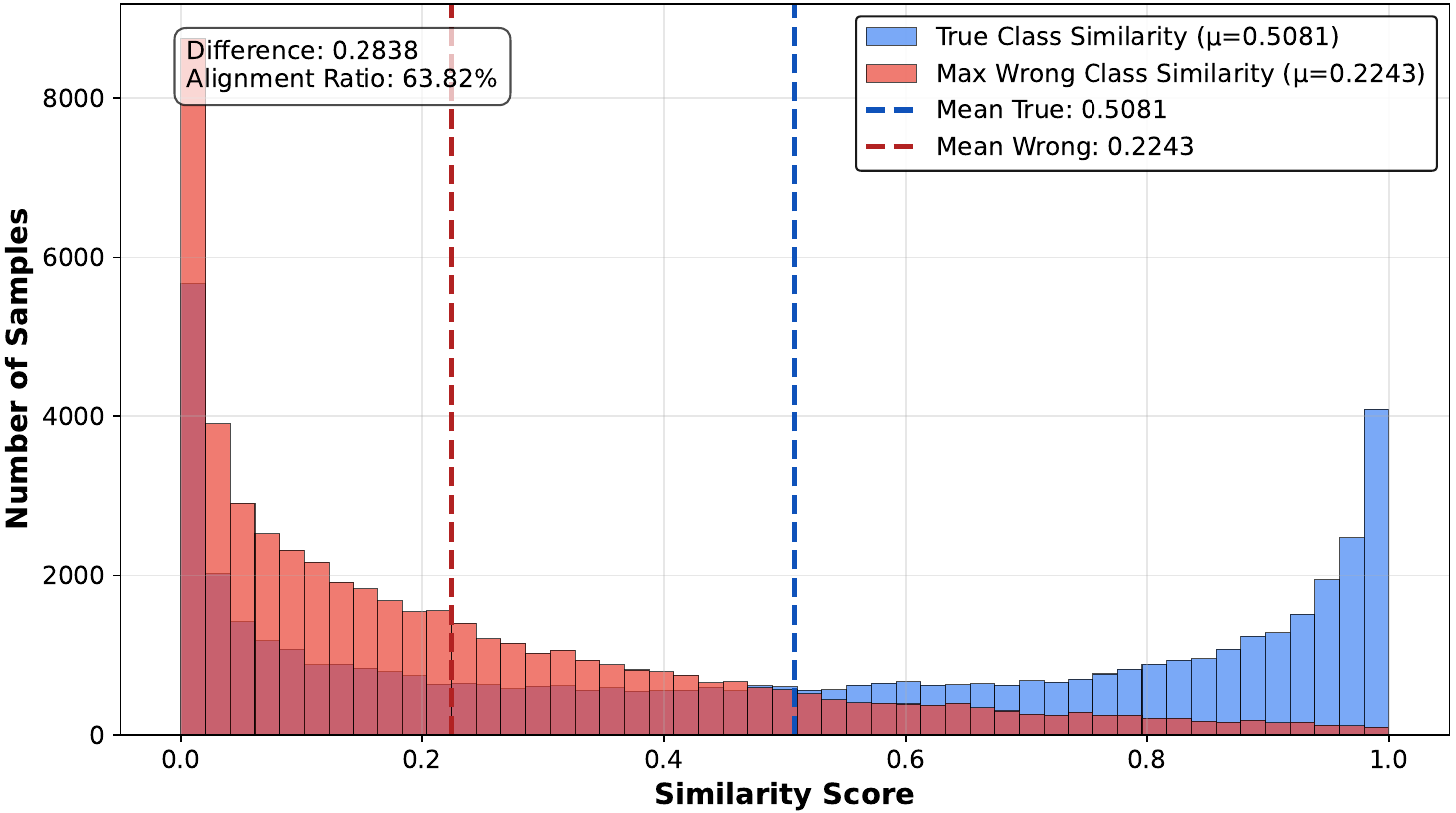} 
\end{center}
\caption{Distribution of True Class Similarity vs. Maximum Wrong Class Similarity for ID images (ImageNet-1k, using CLIP-ViT-B/16). The clear separation between the distributions (blue for true class, red for max wrong class) supports the ID Classification Alignment property.}
\label{fig:appendix_true_vs_wrong_class_sim_hist}
\vspace{-.05in}
\end{figure*}

To provide detailed empirical support, we analyze similarity scores between ID image embeddings (from ImageNet-1k validation set) and text embeddings of all 1000 ImageNet-1k class prompts using CLIP-ViT-B/16.
Figure~\ref{fig:appendix_true_vs_wrong_class_sim_hist} presents histograms of (a) similarity scores to the true class prompt versus (b) the maximum similarity score obtained from any incorrect class prompt, for each ID image. As shown, the distribution of true class similarity scores (mean $\mu \approx 0.51$, based on typical CLIP scores) is significantly shifted towards higher values compared to the distribution of maximum wrong class similarity scores (mean $\mu \approx 0.22$). The substantial mean difference robustly confirms that ID images exhibit a significantly higher affinity to their correct class concept embedding.

\begin{figure*}[h!]
\begin{center}
\includegraphics[width=0.7\textwidth]{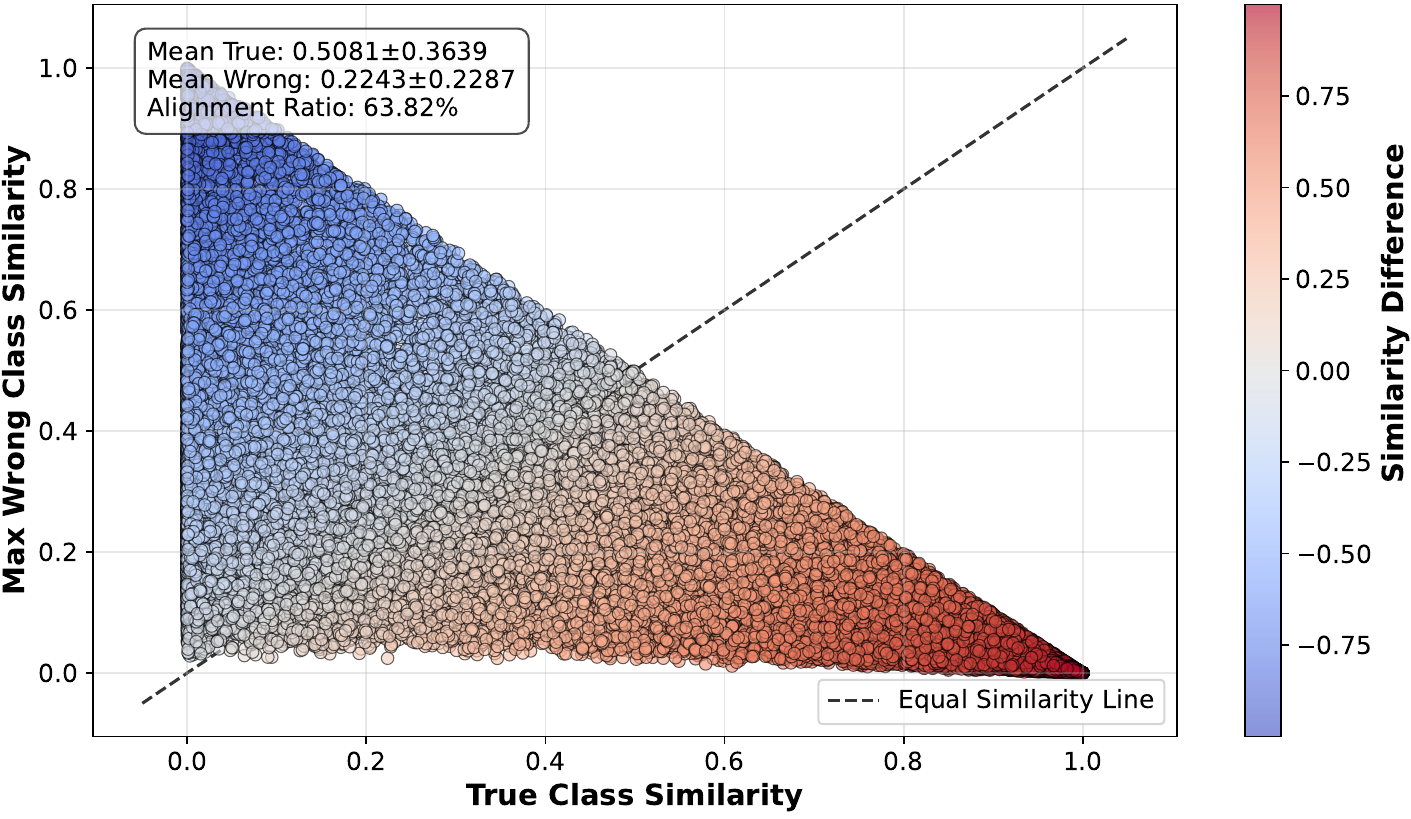} 
\end{center}
\caption{Scatter plot of True Class Similarity vs. Maximum Wrong Class Similarity for ID images (ImageNet-1k, CLIP-ViT-B/16). Each point represents an ID image. The concentration of points above the diagonal ($y=x$) line indicates stronger alignment with the true class prompt.}
\label{fig:appendix_true_vs_wrong_class_scatter}
\vspace{-.05in}
\end{figure*}

Figure~\ref{fig:appendix_true_vs_wrong_class_scatter} provides a scatter plot comparing, for each ID image, its true class similarity against its maximum wrong class similarity. The vast majority of points lie above the $y=x$ "Equal Similarity Line," indicating that true class similarity is generally higher than any incorrect class similarity. The color coding (e.g., by the difference) can further visually reinforce this separation. Only a small fraction of points (typically misclassified samples or ambiguous images) fall below or close to the diagonal, highlighting the strength of this alignment property.

\begin{figure*}[h!]
\begin{center}
\includegraphics[width=0.7\textwidth]{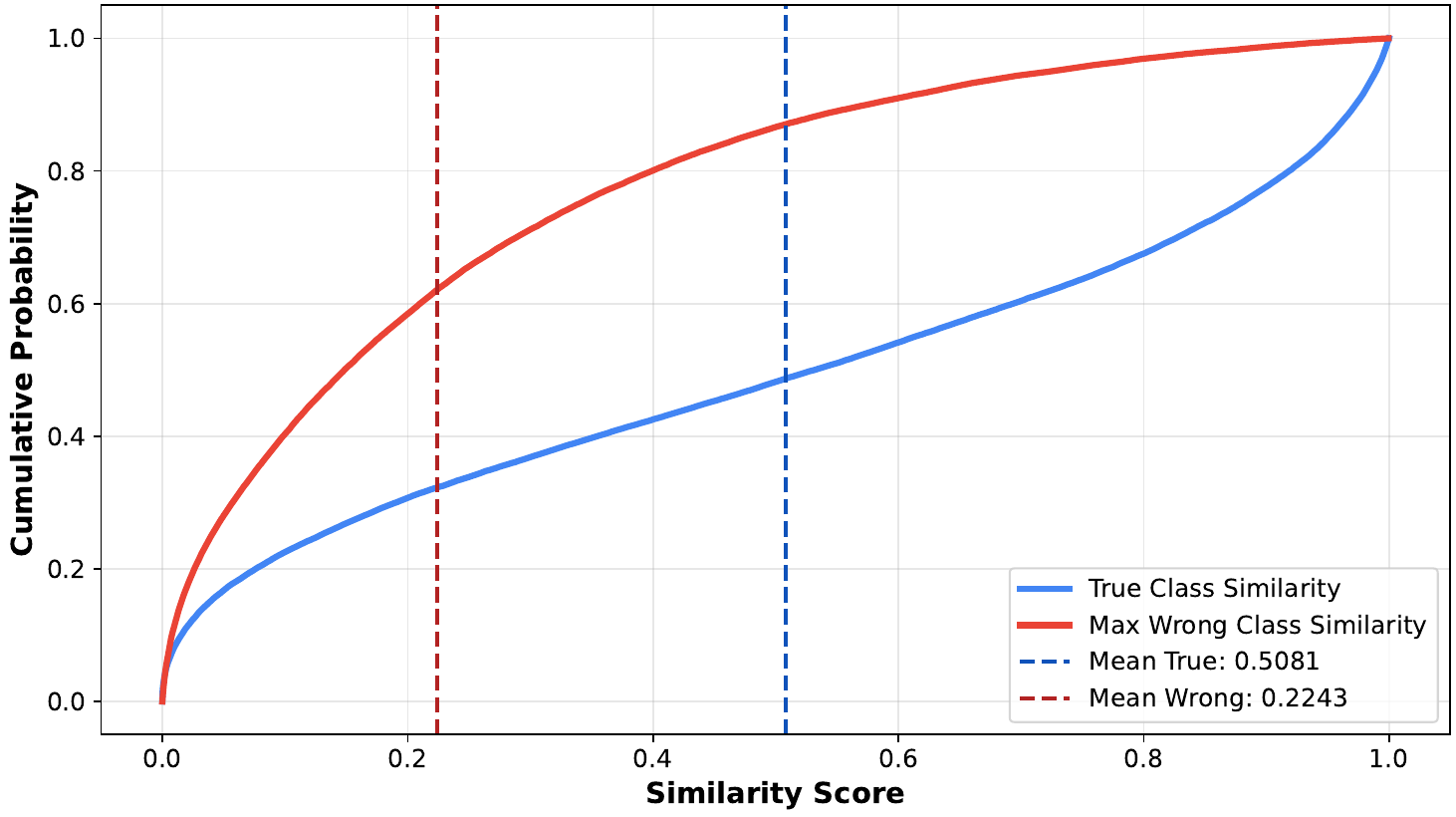} 
\end{center}
\caption{Cumulative Distribution Function (CDF) of True Class Similarity (blue) and Maximum Wrong Class Similarity (red) scores for ID images (ImageNet-1k, CLIP-ViT-B/16). The distinct curves further quantify the ID Classification Alignment.}
\label{fig:appendix_true_vs_wrong_class_cdf}
\vspace{-.05in}
\end{figure*}

Figure~\ref{fig:appendix_true_vs_wrong_class_cdf} shows the Cumulative Distribution Functions (CDFs) for the true class and maximum wrong class similarity scores. The CDF for true class similarity rises much more steeply towards higher similarity values than the CDF for maximum wrong class similarity. This quantitatively demonstrates that for any given similarity threshold, an ID image is significantly more likely to exceed that threshold with its true class prompt than with any incorrect class prompt. These detailed results robustly validate the ID Classification Alignment property and illustrate the semantically discriminative structure learned by the VLM, which is foundational for OOD detection using ID prompts.


\subsubsection{ID vs. OOD Maximum ID Similarity Analysis}
\label{appendix:subsubsec:id_vs_ood_max_sim_appendix} 

Building upon the classification alignment within ID classes, the second property we systematically investigate (Insight~\hyperref[insight1]{1}, Property~\ref{property2-insight1}
) is that the maximum similarity to any ID class prompt ($S_{\text{ID}}(I)$) tends to be significantly higher for ID samples than for OOD samples. This property forms a basic signal for distinguishing ID from OOD based purely on an image's affinity to known concepts.

\begin{figure*}[h!]
\begin{center}
\includegraphics[width=0.7\textwidth]{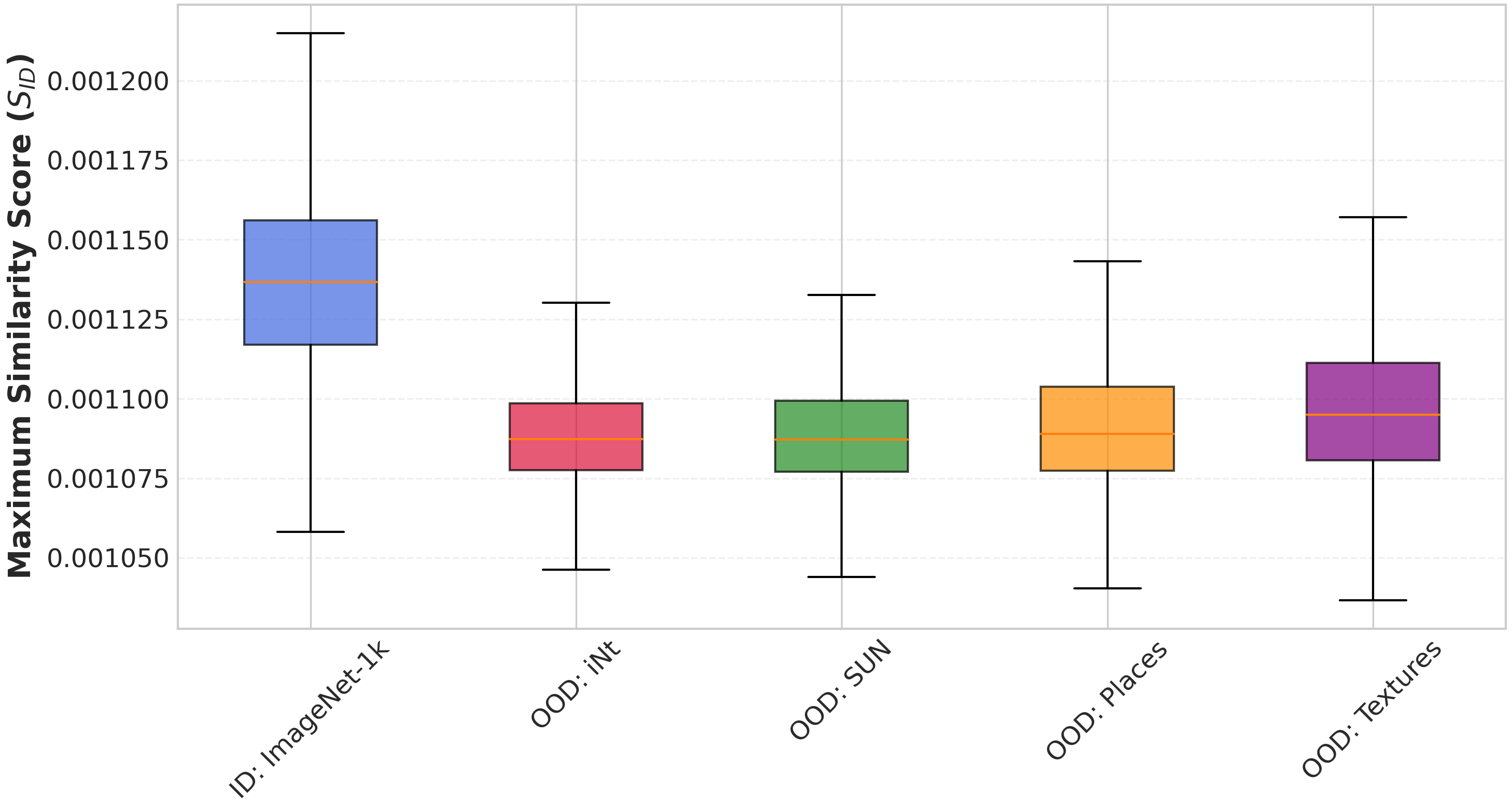} 
\end{center}
\caption{Box plots comparing the distribution of Maximum Similarity to ID Prompts ($S_{\text{ID}}(I)$) for ID (ImageNet-1k) and various OOD datasets (iNaturalist, SUN, Places, Textures), using CLIP-ViT-B/16. The ID distribution is consistently higher, indicating stronger affinity to known concepts for ID samples.}
\label{fig:appendix_sid_distributions_boxplot}
\vspace{-.05in}
\end{figure*}

Figure~\ref{fig:appendix_sid_distributions_boxplot} presents box plots comparing the distribution of $S_{\text{ID}}(I)$ scores for ID images (ImageNet-1k) and images from the four OOD datasets, using CLIP-ViT-B/16. The box plot for ID samples is clearly positioned higher on the similarity axis compared to the box plots for all OOD datasets (iNaturalist, SUN, Places, Textures). This indicates that, on average and across their respective distributions, ID samples exhibit a stronger maximum affinity to the set of known ID concepts than OOD samples do.

\begin{figure*}[h!]
\begin{center}
\includegraphics[width=0.7\textwidth]{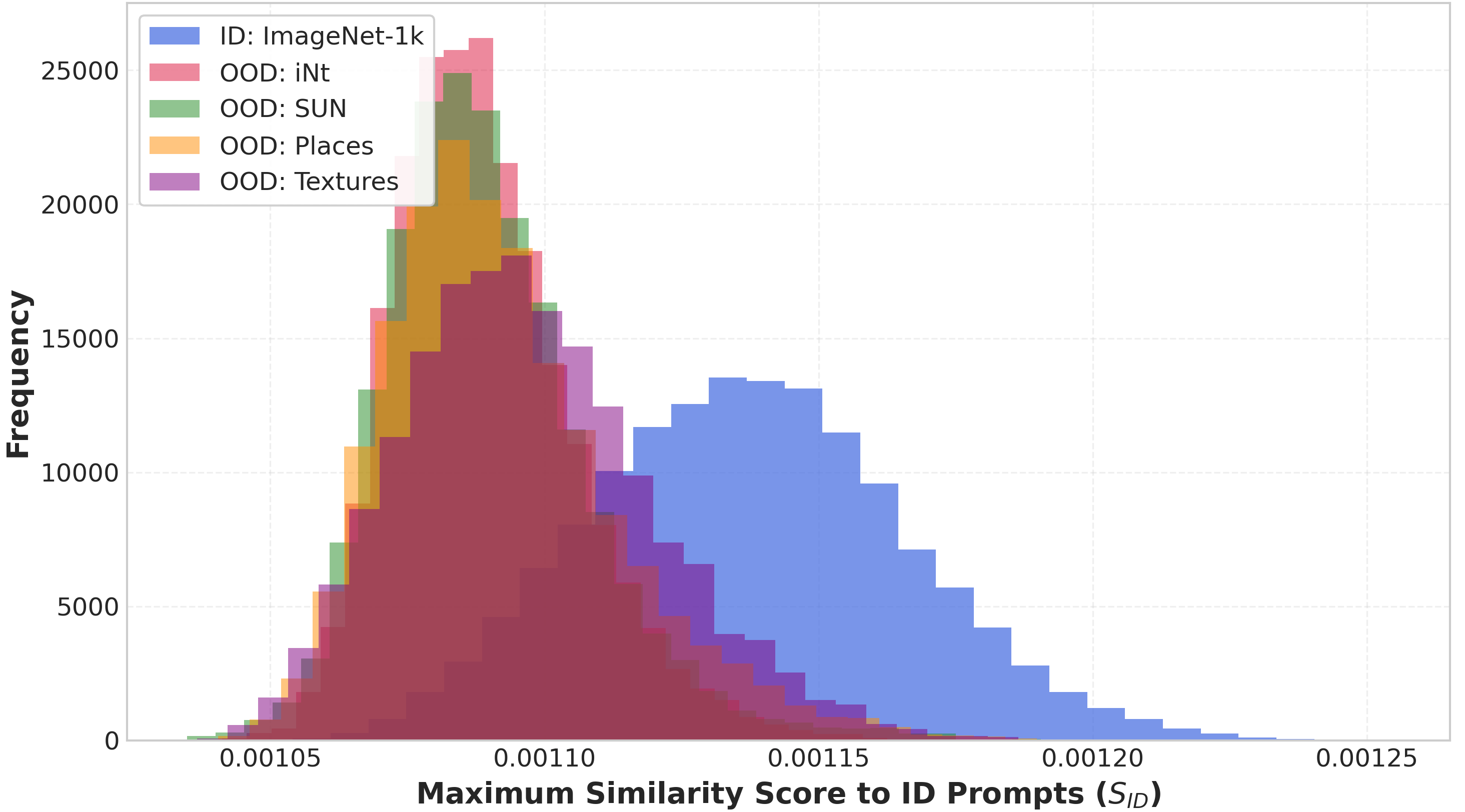} 
\end{center}
\caption{Histograms showing the distribution of Maximum Similarity to ID Prompts ($S_{\text{ID}}(I)$) for ID (ImageNet-1k, blue) and a composite of OOD datasets (other colors), using CLIP-ViT-B/16. The ID distribution is visibly shifted towards higher similarity scores compared to OOD distributions.}
\label{fig:appendix_sid_distributions_hist}
\vspace{-.05in}
\end{figure*}

Figure~\ref{fig:appendix_sid_distributions_hist} further illustrates this with histograms of $S_{\text{ID}}(I)$ scores for ID and the combined OOD datasets. The distribution for ID samples is notably shifted towards higher similarity values, forming a distinct peak at a higher score range. In contrast, the distributions for OOD datasets are generally centered at lower $S_{\text{ID}}(I)$ values, although some overlap exists, particularly for OOD samples that might share some visual or semantic attributes with certain ID classes. This visual separation in the distribution of $S_{\text{ID}}(I)$ scores provides strong empirical evidence for the "ID vs. OOD Maximum ID Similarity Contrast" property. Our analysis further confirms that $S_{\text{ID}}(I)$ serves as a viable, albeit sometimes limited by overlap, signal for OOD detection. The extent of separation in these distributions directly impacts the OOD detection performance achievable when using only $S_{\text{ID}}(I)$ as the OOD score.


\subsubsection{Relative Affinity Scoring Analysis}
\label{appendix:subsubsec:relative_affinity_scoring_appendix} 

The third fundamental property enabling robust VLM-based OOD detection, which our study systematically validates (Insight~\hyperref[insight1]{1}, Property~\ref{property3-insight1}), highlights the enhanced effectiveness achieved by utilizing an image's relative affinity to both in-distribution (ID) and explicitly designed out-of-distribution (OOD) textual prompts. Scores based on this relative affinity, such as the $S_{\mathrm{ID+OOD}}(x)$ score defined in Eq.~\eqref{eq:score_id_ood}, are empirically shown to provide improved separation between ID and OOD samples compared to scores relying solely on affinity to ID concepts (like $S_{\text{ID}}(I)$).

The underlying intuition, confirmed by our experiments, is that ID images will typically exhibit high similarity to their true ID class prompt (Property~\ref{property1-insight1}) and relatively low similarity to well-chosen OOD prompts (which represent concepts semantically distant from ID classes). Conversely, OOD images will generally have lower maximum similarity to ID prompts (Property~\ref{property2-insight1}), but may exhibit comparatively higher similarity to OOD prompts than ID images do, or at least their similarity to ID prompts will be less dominant when OOD prompts are also considered in a normalized scoring scheme. By considering the relationship (e.g., normalized ratio as in Eq.~\eqref{eq:score_id_ood}, or a weighted difference) between these affinities, a more discriminative signal can be derived.

\begin{figure*}[h!]
\centering
\includegraphics[width=\textwidth]{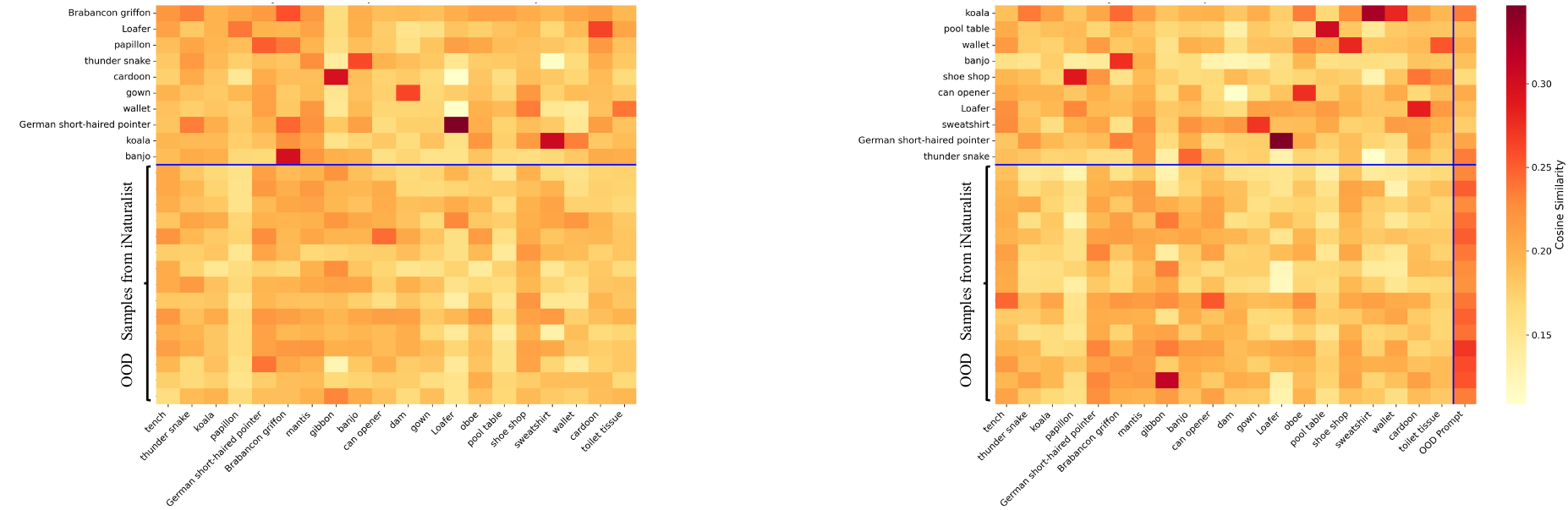}
\caption{Image-Text Cosine Similarity Heatmap for ID (ImageNet-1k) vs. OOD (iNaturalist) samples against ID and OOD prompts (CLIP-ViT-B/16). Demonstrates differential affinities.}
\label{fig:appendix_heatmap_inat}
\end{figure*}

\begin{figure*}[h!]
\centering
\includegraphics[width=\textwidth]{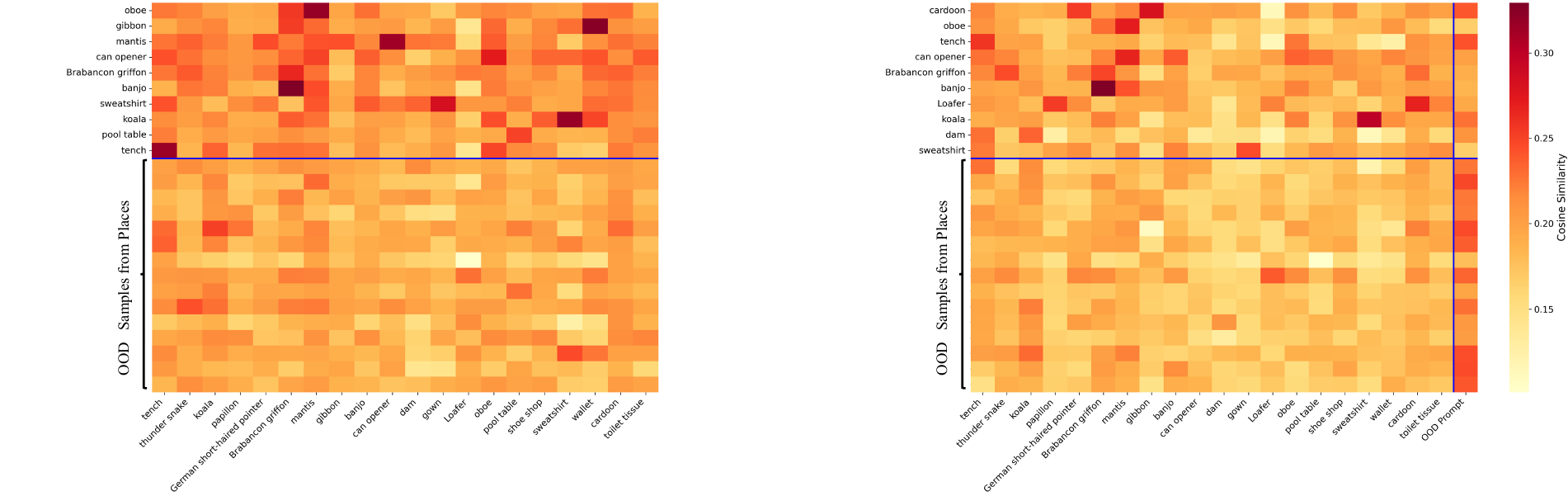}
\caption{Image-Text Cosine Similarity Heatmap for ID (ImageNet-1k) vs. OOD (Places) samples (CLIP-ViT-B/16).}
\label{fig:appendix_heatmap_places}
\end{figure*}

\begin{figure*}[h!]
\centering
\includegraphics[width=\textwidth]{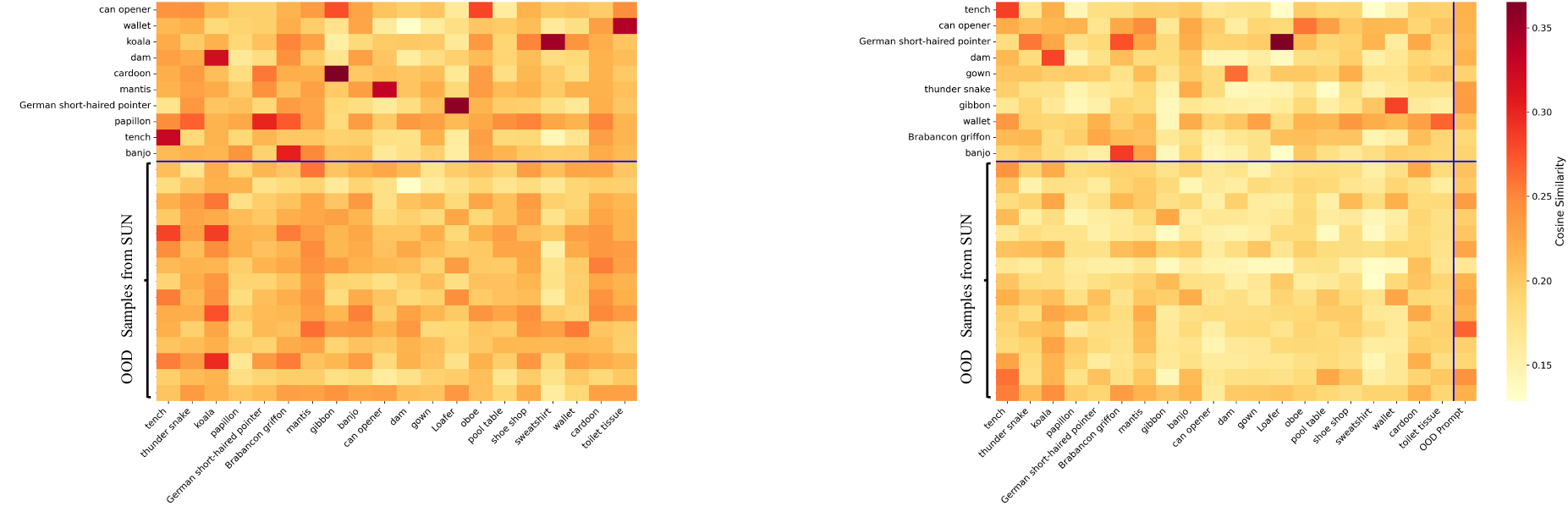}
\caption{Image-Text Cosine Similarity Heatmap for ID (ImageNet-1k) vs. OOD (SUN) samples (CLIP-ViT-B/16).}
\label{fig:appendix_heatmap_sun}
\end{figure*}

\begin{figure*}[h!]
\centering
\includegraphics[width=\textwidth]{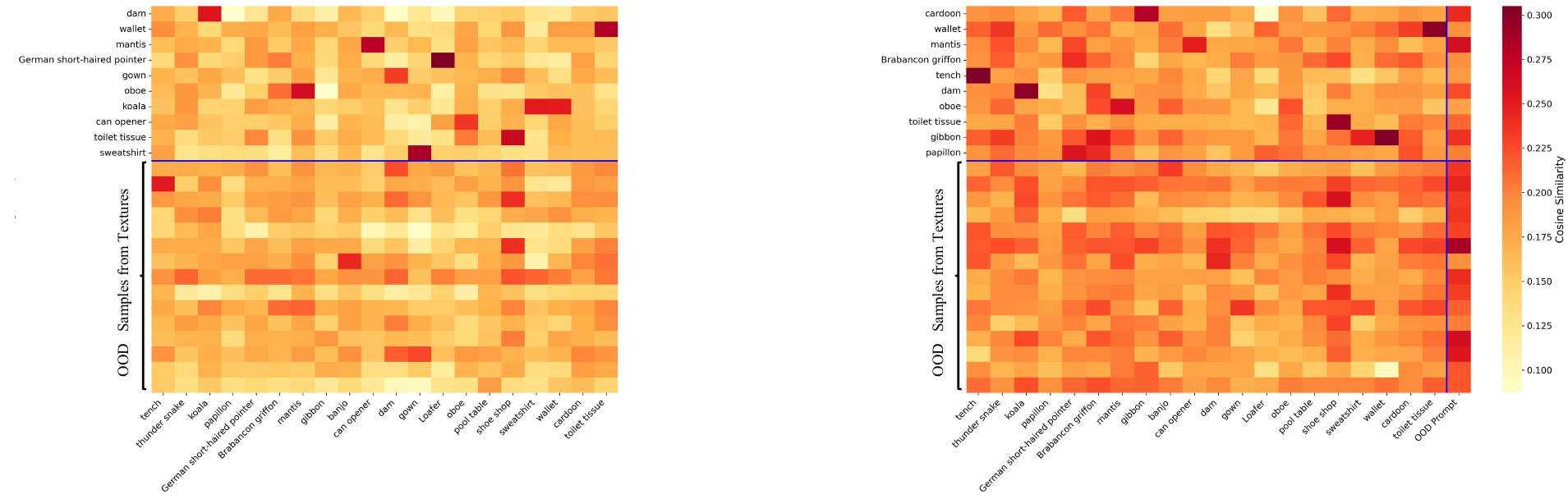}
\caption{Image-Text Cosine Similarity Heatmap for ID (ImageNet-1k) vs. OOD (Textures) samples (CLIP-ViT-B/16).}
\label{fig:appendix_heatmap_textures}
\end{figure*}

Figures~\ref{fig:appendix_heatmap_inat} through~\ref{fig:appendix_heatmap_textures} provide qualitative visual evidence for this relative affinity using cosine similarity heatmaps for each OOD dataset. Each figure typically shows similarities between sample images (ID top rows, OOD bottom rows) and a selection of ID class prompts, alongside an additional column for a representative OOD prompt. For ID images, high similarity is concentrated in relevant ID concept columns, with low similarity to the OOD prompt column. For OOD images, similarity to ID prompts is often lower/more dispersed, while similarity to the OOD prompt column can be notable. This visually confirms that OOD samples often have distinct affinity patterns when OOD prompts are explicitly considered.

\begin{figure*}[h!]
\centering
\includegraphics[width=\textwidth]{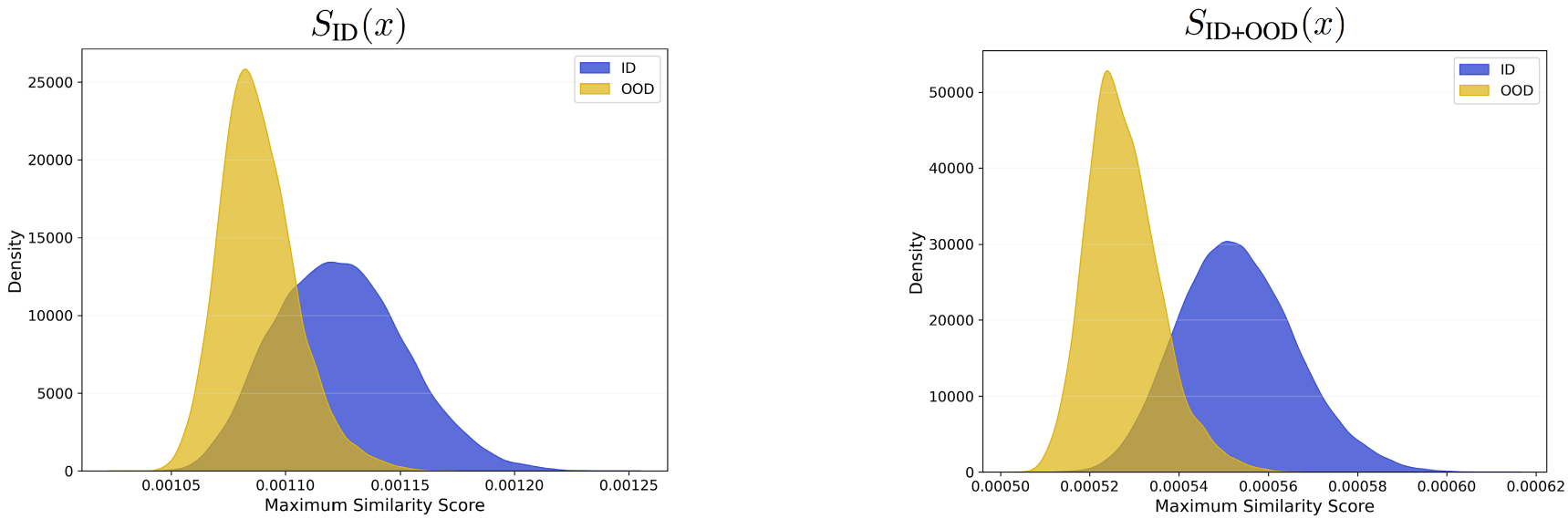}
\caption{Density distributions of OOD scores for ID (ImageNet-1k) vs. OOD (iNaturalist) samples, using CLIP-ViT-B/16. Left: $S_{\text{ID}}(x)$. Right: $S_{\mathrm{ID+OOD}}(x)$.}
\label{fig:appendix_score_dist_inat}
\end{figure*}

\begin{figure*}[h!]
\centering
\includegraphics[width=\textwidth]{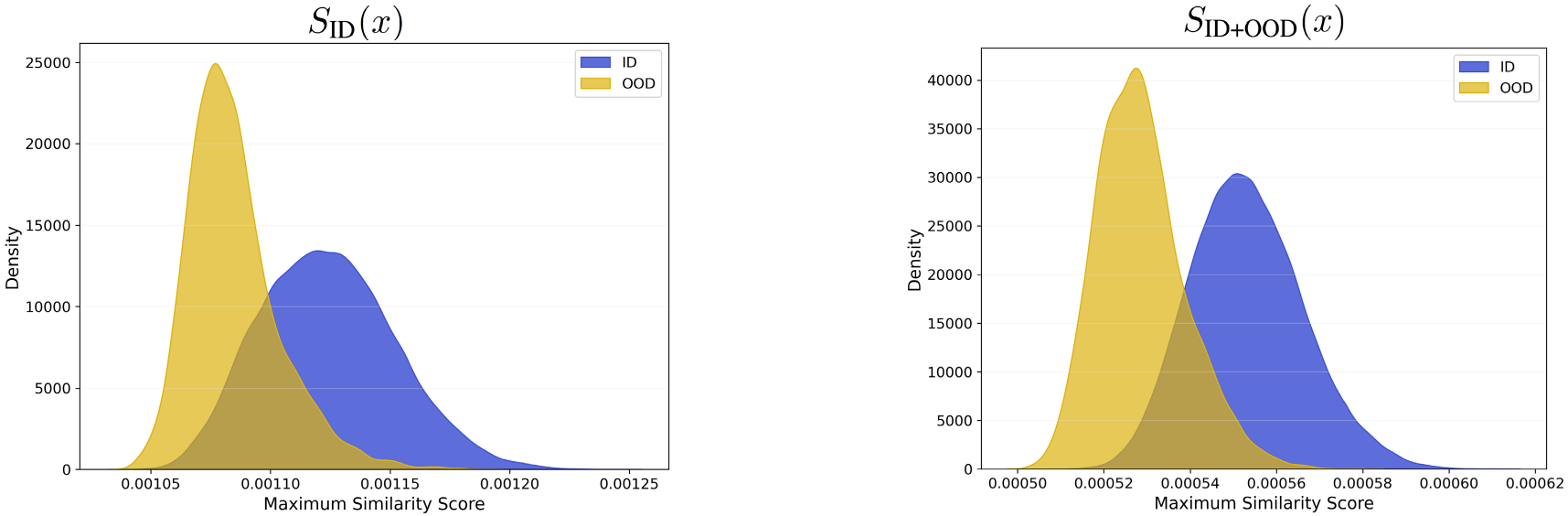}
\caption{Density distributions of OOD scores for ID (ImageNet-1k) vs. OOD (Places) samples (CLIP-ViT-B/16). Left: $S_{\text{ID}}(x)$. Right: $S_{\mathrm{ID+OOD}}(x)$.}
\label{fig:appendix_score_dist_places}
\end{figure*}

\begin{figure*}[h!]
\centering
\includegraphics[width=\textwidth]{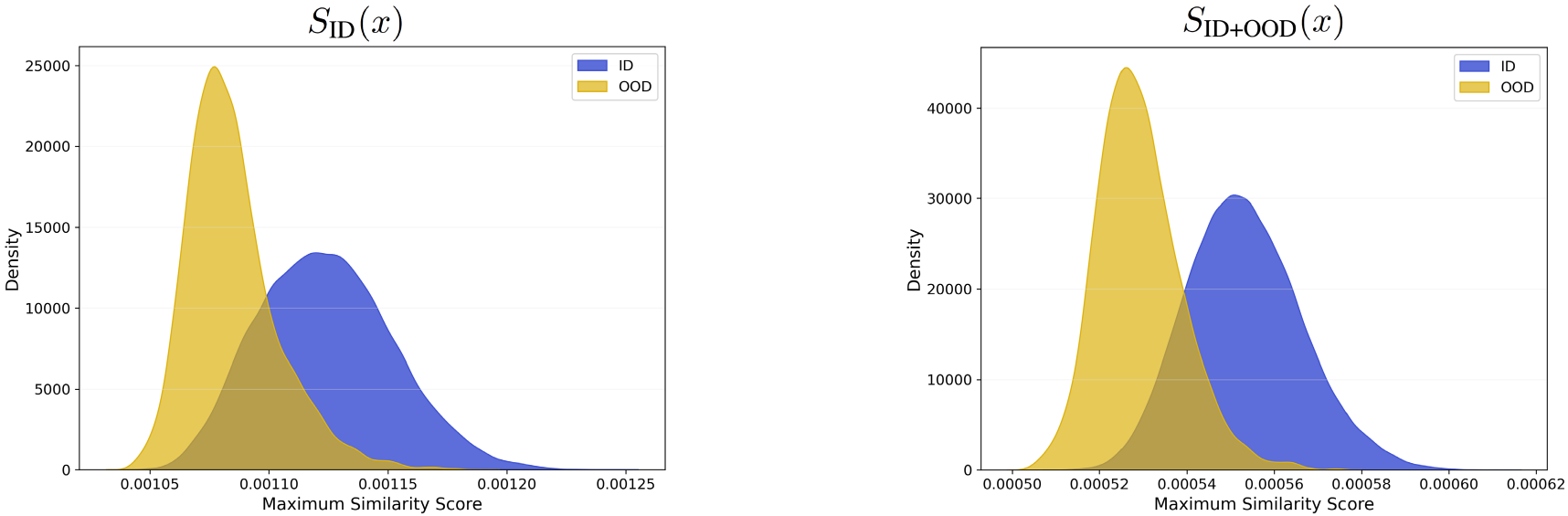}
\caption{Density distributions of OOD scores for ID (ImageNet-1k) vs. OOD (SUN) samples (CLIP-ViT-B/16). Left: $S_{\text{ID}}(x)$. Right: $S_{\mathrm{ID+OOD}}(x)$.}
\label{fig:appendix_score_dist_sun}
\end{figure*}

\begin{figure*}[h!]
\centering
\includegraphics[width=\textwidth]{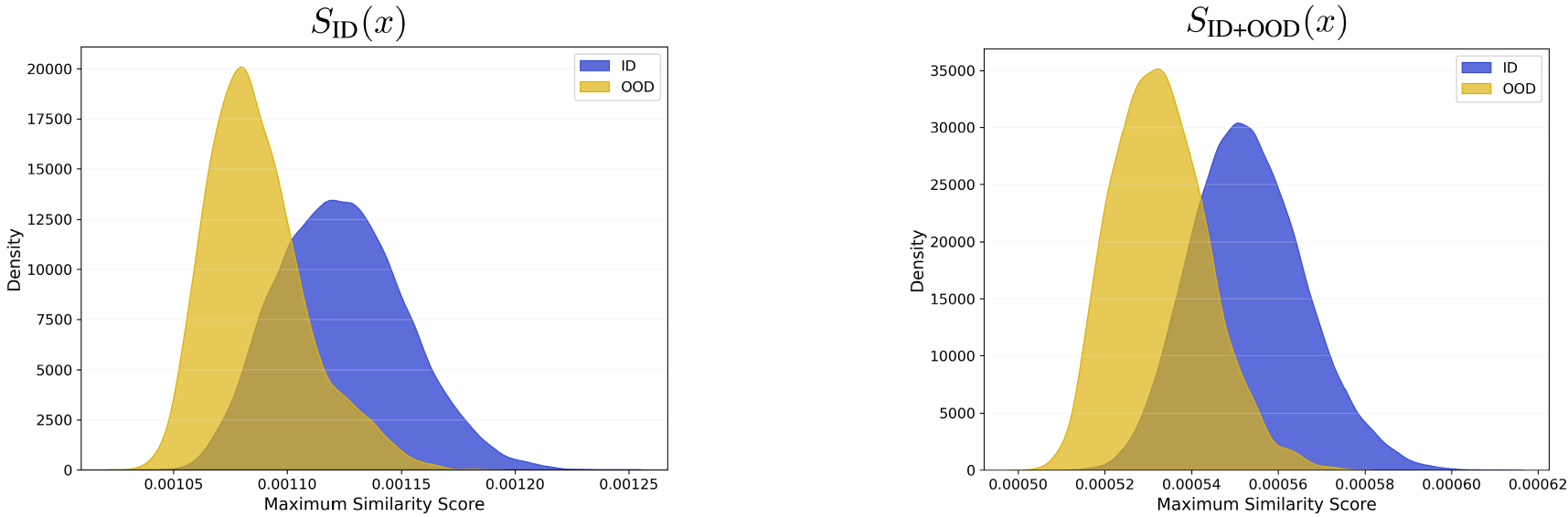}
\caption{Density distributions of OOD scores for ID (ImageNet-1k) vs. OOD (Textures) samples (CLIP-ViT-B/16). Left: $S_{\text{ID}}(x)$. Right: $S_{\mathrm{ID+OOD}}(x)$.}
\label{fig:appendix_score_dist_textures}
\end{figure*}

More quantitatively, Figures~\ref{fig:appendix_score_dist_inat} through~\ref{fig:appendix_score_dist_textures} illustrate the improved separation achieved by using the $S_{\mathrm{ID+OOD}}(x)$ score. Each figure compares the density distributions of scores for ID (e.g., blue) and a specific OOD dataset (e.g., orange/yellow): the left panel shows the distribution using only $S_{\text{ID}}(x)$, while the right panel shows the distribution using $S_{\mathrm{ID+OOD}}(x)$. Across all four OOD datasets, the distributions using the $S_{\mathrm{ID+OOD}}(x)$ score (right panels) consistently show significantly better separation (less overlap, more distant means/modes) between ID and OOD samples compared to using $S_{\text{ID}}(x)$ alone (left panels). It demonstrates how leveraging the image's position relative to both ID and OOD concept references in the VLM space provides a more powerful mechanism for distinguishing between known and novel inputs.




\subsection{Robustness and Sensitivity Analysis Details}
\label{appendix:subsec:robustness_sensitivity_details_appendix} 
This section provides detailed experimental results and additional analysis concerning the robustness and sensitivity of VLM-based OOD detection methods to various input perturbations, complementing the discussion in Section~\ref{sec:analysis} and forming the empirical basis for Insight~\hyperref[insight3]{3}. Understanding these behaviors is critical for assessing the practical reliability and potential failure modes of these methods in real-world deployment scenarios where inputs may not be pristine.

\subsubsection{Robustness to Image Perturbations}
\label{appendix:subsubsec:image_robustness_details_appendix} 

A key aspect of reliable OOD detection is robustness to expected variations or corruptions in the input data. Insight~\hyperref[insight3]{3}, Property~\ref{property1-insight3}, suggests that VLM-based OOD detection is relatively robust to bounded image noise. To investigate this, we evaluate the performance of the VLM-based OOD method (using CLIP-ViT-B/16 and the $S_{\mathrm{ID+OOD}}(x)$ score) under increasing levels of common image corruptions applied to both ID and OOD test samples.

\begin{figure*}[h!]
\begin{center}
\includegraphics[width=0.90\textwidth]{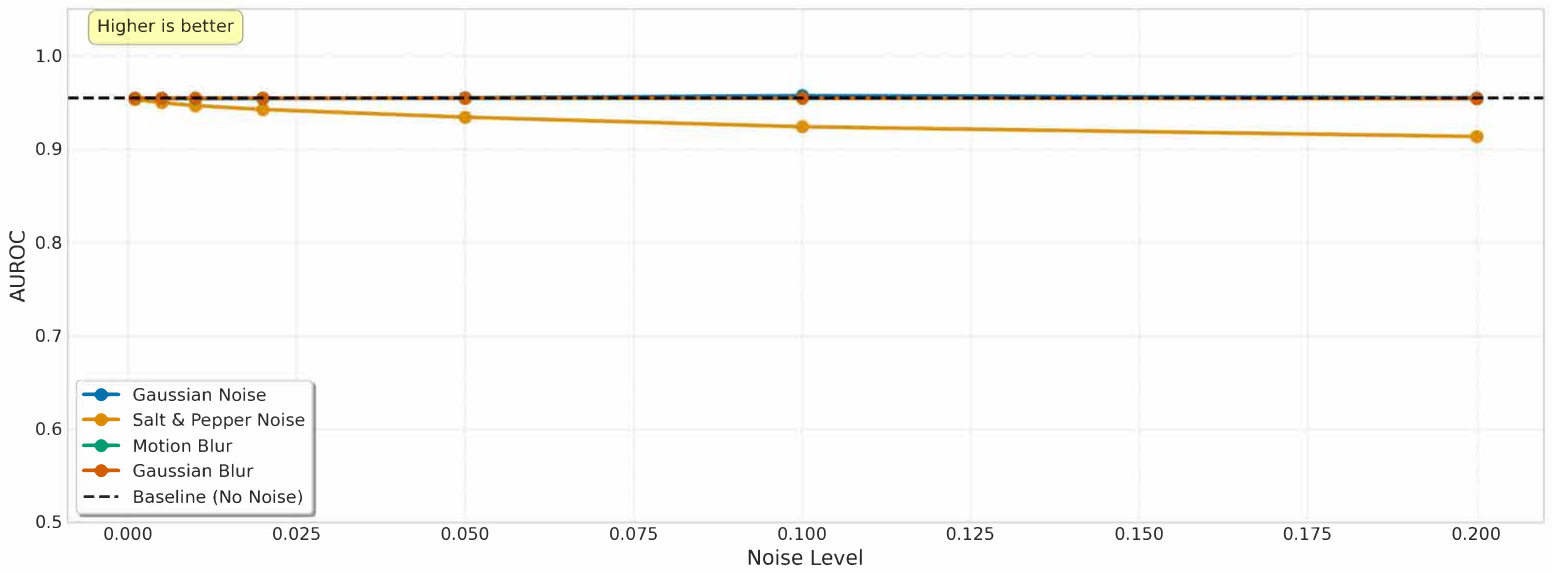} 
\end{center}
\caption{Robustness of VLM-based OOD detection (CLIP-ViT-B/16, AUROC) against increasing severity levels of common image corruptions (types indicated in legend). Performance remains largely stable for many corruptions, demonstrating resilience to image noise and degradations.}
\label{fig:image_robustness_curves_appendix}
\vspace{-.05in}
\end{figure*}

Figure~\ref{fig:image_robustness_curves_appendix} presents the AUROC performance of the VLM-based OOD detector as a function of the corruption intensity level for several types of image noise (e.g., Gaussian, Shot, Impulse) and blur (e.g., Defocus, Motion, Gaussian). The horizontal dashed line represents the baseline performance on clean, uncorrupted images (Severity Level 0). As shown, for most corruption types, the OOD detection performance (AUROC) remains remarkably stable or degrades gracefully, even at higher severity levels. For instance, Gaussian noise and most blurs have a limited impact. Some high-frequency noise like "Salt \& Pepper" (Impulse noise) can cause a more noticeable, albeit still often moderate, decrease in AUROC at high intensities. This empirical evidence strongly supports Insight~\hyperref[insight3]{3} (Property~\ref{property1-insight3}), demonstrating that the VLM-based OOD detection method, by leveraging high-level semantic features, is relatively resilient to common pixel-level noise and visual degradations. This aligns with the known robustness characteristics of VLMs like CLIP for standard classification tasks \cite{CLIP}.

\subsubsection{Sensitivity to Text Prompt Variations}
\label{appendix:subsubsec:text_prompt_variations_appendix} 

Unlike single-modal methods, VLM-based OOD detection critically relies on textual prompts to define both ID and OOD concepts. The precise wording, phrasing, and structure of these prompts, even if intended to convey the same underlying semantic concept, can potentially influence performance significantly. Insight~\hyperref[insight3]{3} (Property~\ref{property2-insight3}) posits such a sensitivity. To explore this, we evaluate our VLM-based OOD method (CLIP-ViT-B/16, $S_{\mathrm{ID+OOD}}(x)$ score) using sets of ID prompts that are deliberate variations of a baseline ID prompt set (e.g., using synonyms for prompts or class labels; paraphrasing entire ID prompts; altering sentence structure). OOD prompts are kept consistent. We measure the resulting change in OOD detection performance (AUROC) compared to using the baseline ID prompt set.

\begin{figure*}[h!]
\begin{center}
\includegraphics[width=0.80\textwidth]{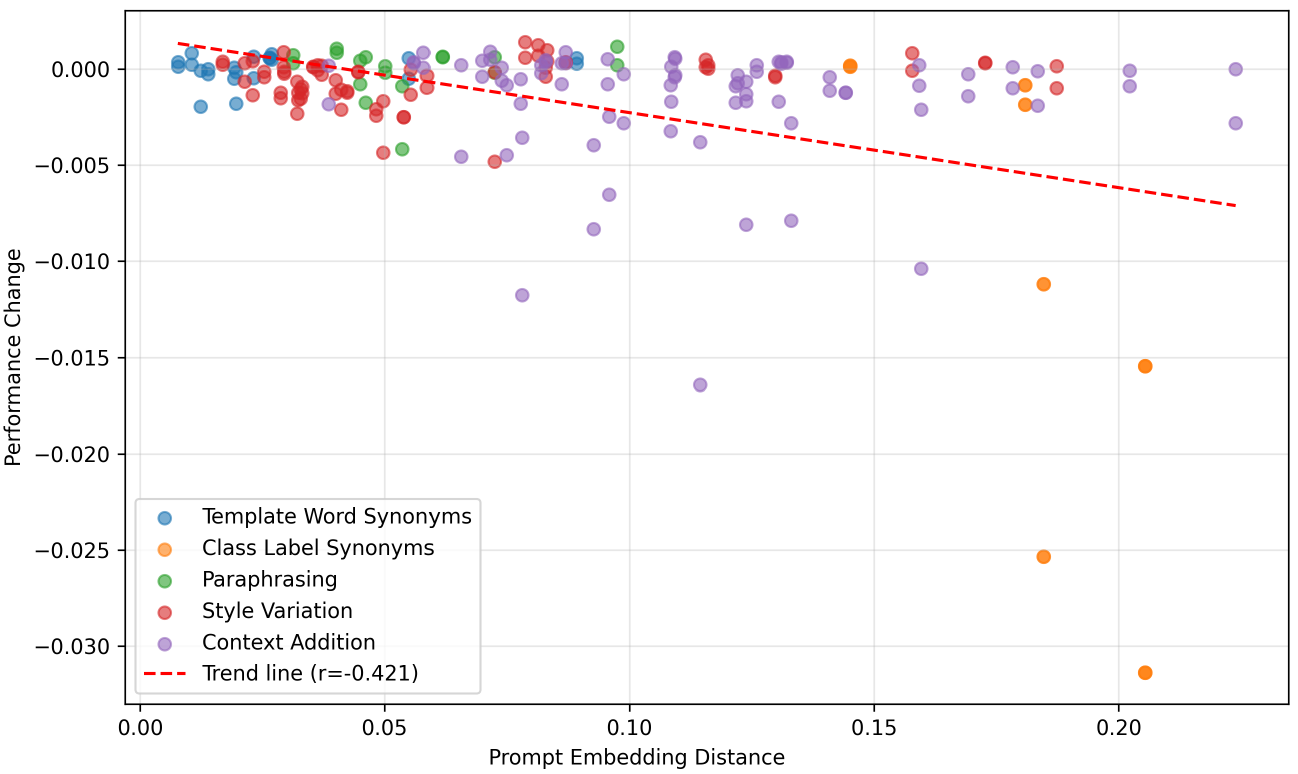} 
\end{center}
\caption{Sensitivity of VLM-based OOD detection performance (change in AUROC using CLIP-ViT-B/16) to variations in ID text prompts. The x-axis represents the embedding distance between the varied ID prompt set and a baseline set. The y-axis shows the $\Delta$AUROC. The plot indicates that changes in prompt wording, leading to different text embeddings, are likely to result in changes (often degradation) in OOD detection performance.}
\label{fig:prompt_sensitivity_scatter_appendix}
\vspace{-.05in}
\end{figure*}

Figure~\ref{fig:prompt_sensitivity_scatter_appendix} displays a scatter plot illustrating the relationship between the "Prompt Embedding Distance" (e.g., average cosine distance in the CLIP text embedding space between embeddings of the varied ID prompt set and the baseline ID prompt set) and the corresponding change in OOD detection AUROC. Each point represents the performance change for a specific set of varied ID prompts. The plot reveals a general negative trend (correlation coefficient $r \approx -0.42$ in the example shown): as the embedding distance of the varied ID prompts from the baseline increases (implying greater semantic or structural difference as perceived by the VLM), the OOD detection performance tends to decrease, sometimes substantially. This empirical relationship confirms Insight~\hyperref[insight3]{3} (Property~\ref{property2-insight3}), highlighting that VLM-based OOD detection performance is indeed sensitive to the specific wording and semantic positioning of the text prompts, especially ID prompts. This underscores that prompt engineering is a non-trivial and critical factor for achieving optimal and consistent results with these methods.

\subsubsection{Vulnerability to Text Perturbations}
\label{appendix:subsubsec:text_perturbations_details_appendix} 

Beyond deliberate variations in prompt design, textual inputs can also be subjected to other forms of perturbations, including accidental errors (e.g., typos) or, more concerningly, subtle manipulations that might be designed to be human-imperceptible yet disrupt model processing. Insight~\hyperref[insight3]{3} (Property~\ref{property3-insight3}) suggests that VLM-based OOD detection might be particularly vulnerable to such text perturbations, exhibiting an asymmetry compared to its relative image robustness.

\begin{figure*}[h!]
\begin{center}
\includegraphics[width=\textwidth]{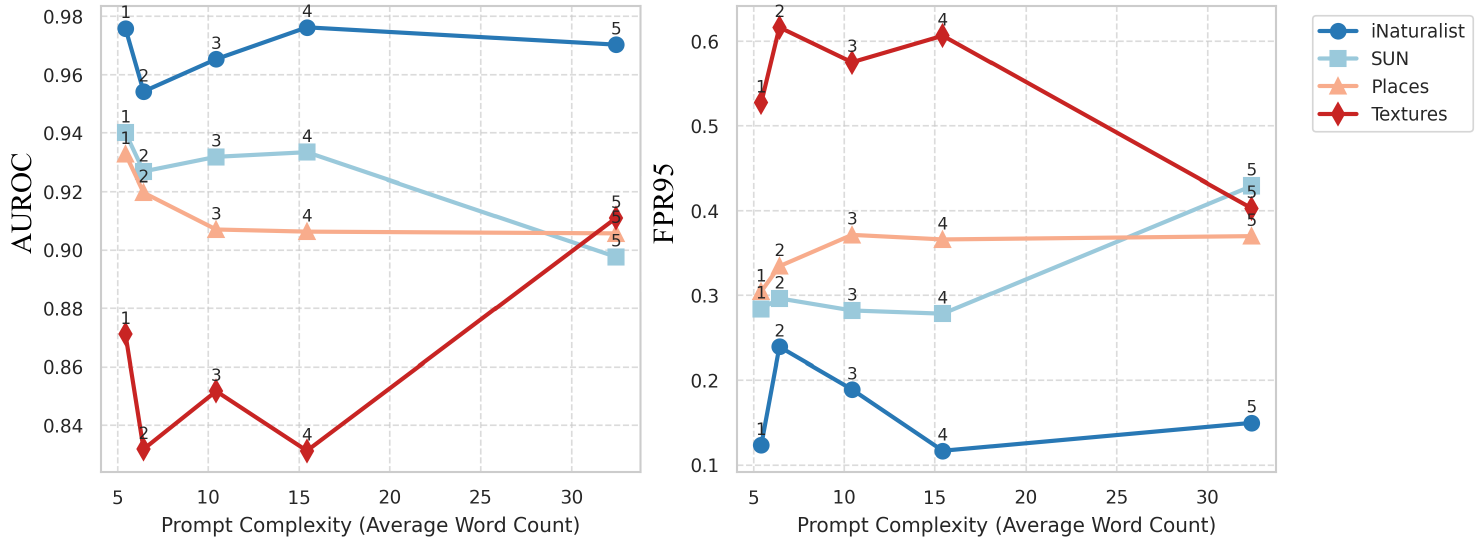} 
\end{center}
\caption{Impact of OOD Prompt Complexity (measured by average word count in OOD prompts) on OOD detection performance (CLIP-ViT-B/16) across different OOD datasets.}
\label{fig:appendix_prompt_complexity_word_count}
\vspace{-.05in}
\end{figure*}

\begin{figure*}[h!]
\begin{center}
\includegraphics[width=\textwidth]{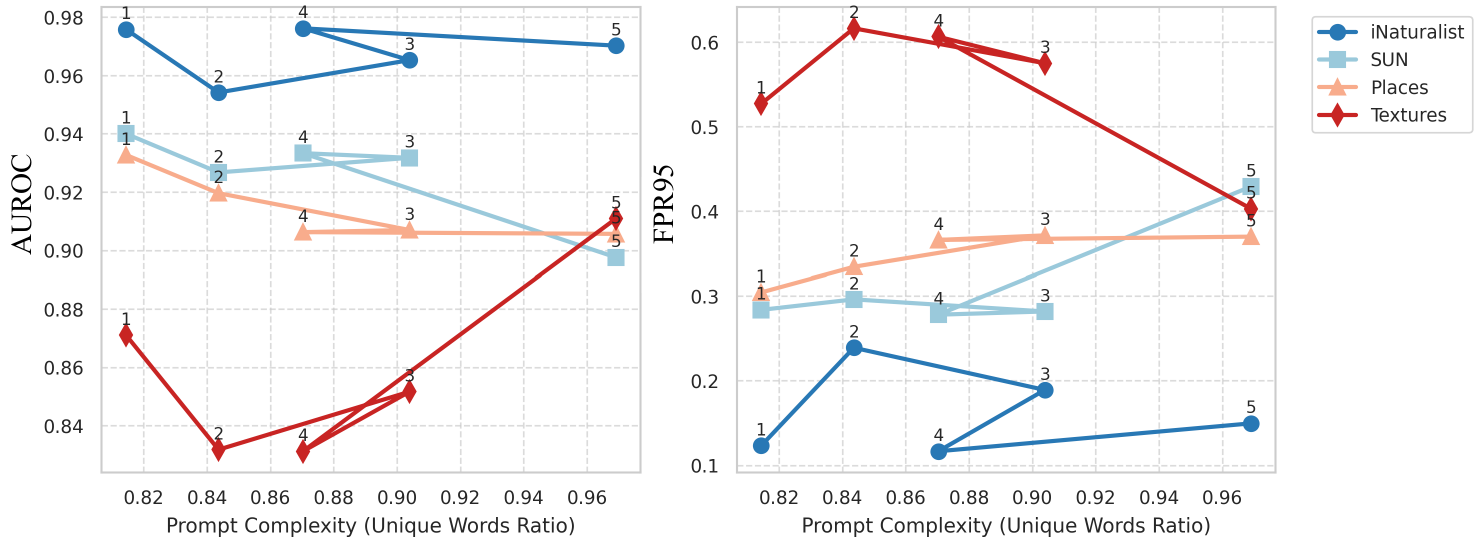} 
\end{center}
\caption{Impact of OOD Prompt Complexity (measured by unique words ratio in OOD prompts) on OOD detection performance (CLIP-ViT-B/16) across different OOD datasets.}
\label{fig:appendix_prompt_complexity_unique_ratio}
\vspace{-.05in}
\end{figure*}

To investigate this, we analyze the impact of prompt complexity variations (as a form of structural perturbation).Figures~\ref{fig:appendix_prompt_complexity_word_count} and \ref{fig:appendix_prompt_complexity_unique_ratio} explore the impact of OOD prompt complexity on OOD detection performance (AUROC and FPR95 with CLIP-ViT-B/16) across different OOD datasets. Complexity is measured by average word count of OOD prompts and the ratio of unique words in them. While not direct adversarial attacks, these experiments reveal the sensitivity of performance to structural changes in prompts. For example, Figure~\ref{fig:appendix_prompt_complexity_word_count} shows that performance can fluctuate significantly as the average word count of OOD prompts changes, with different OOD datasets showing varied sensitivity patterns (e.g., iNaturalist performance peaks with moderately complex OOD prompts, while Textures performance can degrade with increased OOD prompt length). Similarly, Figure~\ref{fig:appendix_prompt_complexity_unique_ratio} shows that the lexical diversity (unique word ratio) of OOD prompts can also influence performance, sometimes non-monotonically.

Collectively, these findings strongly suggest that the textual domain represents a key fragility point for VLM-based OOD detection, necessitating careful prompt design, validation, and potentially the development of text-based adversarial defenses or robustification techniques.

\subsection{Ablation Studies and Method Analysis}
\label{appendix:subsec:ablation_studies_appendix} 

This section presents additional experimental analyses and ablation studies designed to further investigate specific factors influencing the performance and behavior of VLM-based OOD detection methods, particularly focusing on the framework utilizing ID and OOD prompts as described in Appendix~\ref{appendix:subsec:algorithm}. These studies help to understand the sensitivity to hyperparameters and design choices.

\subsubsection{Sensitivity to Temperature Parameter $\tau$ in Scoring Function}
\label{appendix:subsubsec:temperature_sensitivity_appendix} 

The $S_{\mathrm{ID+OOD}}(x)$ scoring function (Equation~\ref{eq:score_id_ood}) includes a temperature parameter $\tau$, which controls the sharpness of the softmax-like normalization of similarity scores. The choice of $\tau$ can significantly impact the final OOD score distribution and thus OOD detection performance. To analyze this sensitivity, we evaluate the average OOD detection performance (AUROC and FPR95, averaged across the four OOD datasets) of our VLM-based method using various CLIP backbones with different values of $\tau$, ranging typically from $0.001$ to $2.0$.

\begin{figure*}[h!]
\begin{center}
\includegraphics[width=\textwidth]{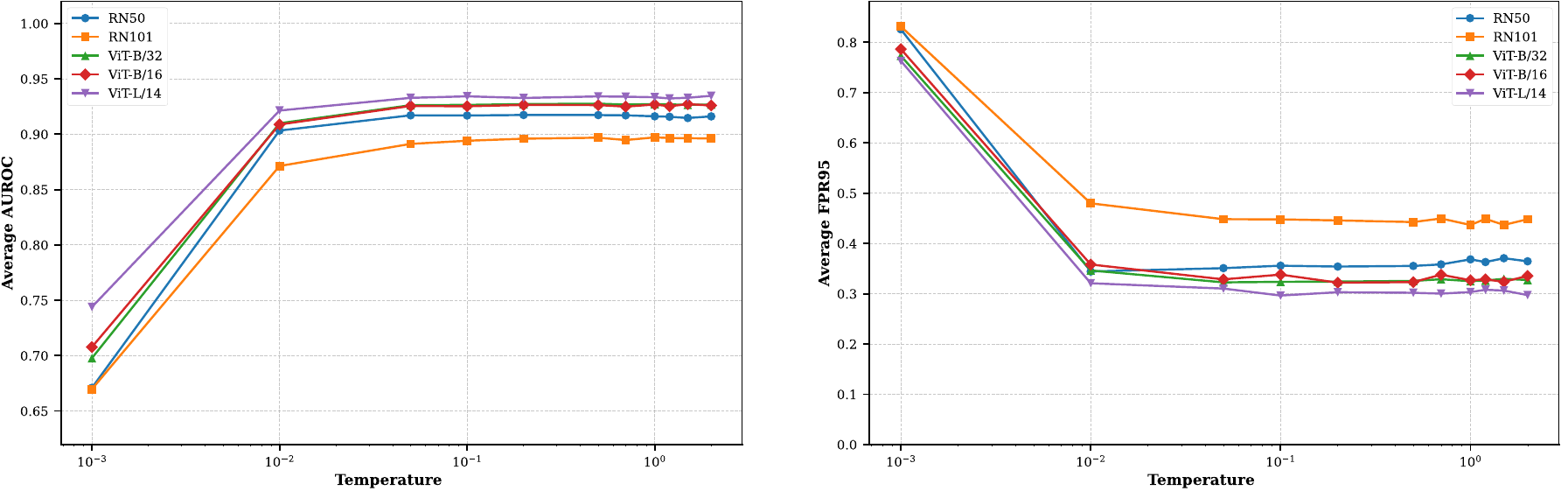} 
\end{center}
\caption{Sensitivity of average OOD detection performance (AUROC and FPR95, averaged over iNaturalist, SUN, Places, Textures) to the temperature parameter $\tau$ in the $S_{\mathrm{ID+OOD}}(x)$ scoring function, shown for different VLM backbones.}
\label{fig:appendix_temperature_sensitivity_plot}
\vspace{-.05in}
\end{figure*}

Figure~\ref{fig:appendix_temperature_sensitivity_plot} plots the average AUROC and FPR95 as $\tau$ varies (shown on a logarithmic scale for clarity). As shown, OOD detection performance is indeed not sensitive to the temperature parameter within a certain range. There exists a clear increasing range for $\tau$, and empirically, for many CLIP variants, performance stabilizes when $\tau$ increases up to 0.1. The figure also presents performance trends for different VLM backbones (RN50, RN101, ViT-B/32, ViT-B/16, ViT-L/14), indicating that while the exact optimal $\tau$ may slightly vary, the characteristics regarding $\tau$ are relatively consistent across different VLM sizes or architectures, although peak performance values naturally differ.

\subsubsection{Impact of Hierarchical Semantic Concepts in OOD Prompts}
\label{appendix:subsubsec:hierarchical_semantic_alignment_appendix} 

To better understand how the VLM embedding space organizes semantic information at different levels of granularity and how this affects OOD detection when using explicit OOD prompts, we examine the impact of incorporating higher-level semantic concepts (which might be superclasses of ID concepts) as part of the OOD prompt set. Specifically, we used specialized, fine-grained category datasets as ID (Oxford-IIIT Pets for "pet" concepts, Food-101 for "food" items, CUB-200 for "bird" species, Stanford Dogs for "dog" breeds). Then, alongside our standard generic OOD prompts (like "an unrelated object"), we introduced additional OOD prompts corresponding to the broad, higher-level category names of the ID dataset itself (e.g., using the prompt "a photo of a pet" as an OOD prompt when Oxford-IIIT Pets is the ID dataset). This tests if including such broad, potentially semantically encompassing concepts as OOD prompts influences the model's ability to distinguish finer-grained OOD samples from the specific ID categories.

\begin{figure*}[h!]
\begin{center}
\includegraphics[width=\textwidth]{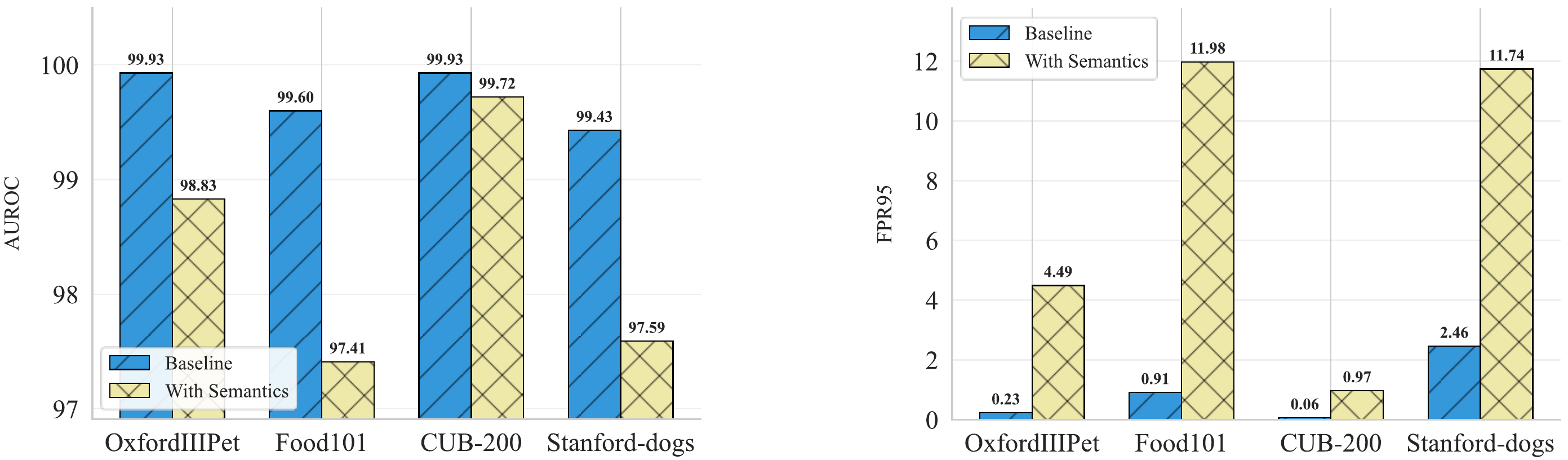} 
\end{center}
\caption{Performance comparison between the baseline VLM-based OOD method and a variant incorporating higher-level semantic concepts that are superclasses of the ID data (e.g., using "pet" as an OOD prompt when ID is OxfordIIIPet) as additional OOD prompts. Evaluated on four specialized ID datasets (Pets, Food101, CUB-200, Stanford Dogs). Incorporating such hierarchical semantics as OOD prompts significantly degrades performance.}
\label{fig:appendix_hierarchical_semantics_effect}
\vspace{-.05in}
\end{figure*}

Figure~\ref{fig:appendix_hierarchical_semantics_effect} compares the OOD detection performance (AUROC and FPR95) between the baseline method (using standard ID prompts and generic OOD prompts) and the variant method incorporating these higher-level semantic concepts as additional OOD prompts, evaluated across the four specialized ID datasets. The results clearly show that incorporating these hierarchical, semantically encompassing concepts as OOD prompts \textit{significantly degrades} OOD detection performance compared to the baseline method for all tested fine-grained ID datasets. Both AUROC drops considerably, and FPR95 increases substantially.

This counter-intuitive finding suggests that while VLMs do capture hierarchical semantics to some extent, forcing the OOD prompts to include broad, higher-level concepts that are effectively superclasses of the ID concepts (e.g., using "pet" as an OOD prompt when detecting OOD from specific pet breeds) can confuse or dilute the relative affinity scoring mechanism (Eq.~\eqref{eq:score_id_ood}). It appears that the relative affinity scoring works best when ID prompts define specific concepts and OOD prompts define concepts that are clearly semantically outside or unrelated to the ID concept space, rather than broader supercategories that the ID concepts themselves belong to. This points to a potential limitation or a nuanced requirement in how the relative affinity scoring handles complex or hierarchical relationships between ID and OOD prompts, and underscores the importance of careful OOD prompt design to avoid concepts that are too semantically close to, or are superclasses of, the ID classes. These prompts should ideally represent "otherness" effectively.

\subsubsection{Choice of VLM Backbone}
\label{appendix:subsubsec:vlm_backbone_appendix} 

The performance of any VLM-based OOD detection method is inherently dependent on the underlying VLM's architecture, its pre-training dataset, and the quality of its learned joint embedding space. Different VLM backbones can therefore lead to variations in OOD detection performance even when using the exact same prompting strategy and scoring function. We compare the OOD detection performance using our unified ID/OOD prompt scoring framework (Algorithm~\ref{alg:Unified_OOD_Detection_Pipeline}) across a range of different publicly available pre-trained VLM backbones.

Table~\ref{table:appendix_vlm_backbone_comparison} presents the OOD detection performance for various CLIP \cite{CLIP} and OpenCLIP \cite{openclip} models (with different vision backbones like ResNet, ViT-B, ViT-L, and trained on different datasets like WebImageText or LAION), as well as ALIGN-Base \cite{ALIGN}. The results show considerable variation in performance depending on the VLM backbone chosen. Generally, larger or more recent ViT-based CLIP/OpenCLIP models (e.g., CLIP-ViT-L/14, OpenCLIP-ViT-B/32 pre-trained on larger datasets) tend to achieve better average OOD detection performance than older or ResNet-based CLIP models, likely due to their stronger visual and semantic representation capabilities. However, performance on individual OOD datasets can show nuances; for instance, some OpenCLIP models, while very strong on semantic OODs like iNaturalist, might show slightly lower performance on textural OODs like Textures compared to certain OpenAI CLIP models, possibly reflecting differences in pre-training data characteristics or architectural nuances influencing their representation of texture versus overt semantic content. ALIGN-Base also shows strong performance, particularly on iNaturalist and Places, but exhibits more struggles on Textures.

This comparison highlights that the choice of VLM backbone is a significant factor in achieving optimal OOD detection performance. While the general mechanism of leveraging ID/OOD prompts within the VLM's semantic space applies across these models, the specific quality, structure, and biases of each VLM's joint embedding space—learned during its unique pre-training—directly impact the effectiveness of this OOD detection paradigm. Future work could explore adaptive prompting strategies or fine-tuning techniques tailored to different VLM backbones for optimal OOD performance, though this study focuses on the zero-shot application.

\begin{table*}[h!]
\centering
\resizebox{\textwidth}{!}{
    \begin{tabular}{@{}l *{10}{c} @{}}
        \toprule
        & \multicolumn{2}{c}{iNaturalist} & \multicolumn{2}{c}{SUN} & \multicolumn{2}{c}{Places} & \multicolumn{2}{c}{Textures} & \multicolumn{2}{c}{\textbf{Average}} \\
        \cmidrule(lr){2-3} \cmidrule(lr){4-5} \cmidrule(lr){6-7} \cmidrule(lr){8-9} \cmidrule(lr){10-11}
        \textbf{VLM Backbone} & \scriptsize FPR95$\downarrow$ & \scriptsize AUROC$\uparrow$ & \scriptsize FPR95$\downarrow$ & \scriptsize AUROC$\uparrow$ & \scriptsize FPR95$\downarrow$ & \scriptsize AUROC$\uparrow$ & \scriptsize FPR95$\downarrow$ & \scriptsize AUROC$\uparrow$ & \scriptsize FPR95$\downarrow$ & \scriptsize AUROC$\uparrow$  \\
        \midrule
        CLIP-RN50 (OpenAI) & 24.05 & 95.26 & 37.54 & 91.32 & 44.51 & 88.92 & 39.24 & 90.85 & 36.33 & 91.58  \\
        CLIP-ViT-B/16 (OpenAI)    & 23.80 & 95.46 & 30.52 & 93.40 & 29.82 & 93.18 & 47.38 & 88.48 & 32.87 & 92.62 \\
        CLIP-ViT-L/14 (OpenAI) & 19.52 & 96.34 & 25.78 & 94.69 & 26.54 & 94.29 & 49.48 & 88.21 & 30.33 & 93.38 \\
        OpenCLIP-ViT-B/32 (LAION-400M) & 10.39 & 97.69 & 36.89 & 90.28 & 31.10 & 92.01 & 60.04 & 83.95 & 34.61 & 90.98 \\
        OpenCLIP-ViT-B/16 (LAION-2B) & 18.30 & 95.99 & 38.21 & 90.63 & 33.07 & 91.59 & 55.43 & 85.14 & 36.25 & 90.84 \\
        ALIGN-Base (Google) & 4.58 & 98.95 & 23.33 & 93.39 & 28.57 & 92.04 & 60.74 & 82.12 & 29.31 & 91.63 \\
        \bottomrule
    \end{tabular}
} 
\caption{Performance comparison of various VLM backbones using our ID/OOD prompt framework on different OOD datasets, with ImageNet-1k as the in-distribution. Performance varies significantly with the choice of VLM, highlighting its importance.}
\label{table:appendix_vlm_backbone_comparison}
\end{table*}

\section{Limitations}
\label{appendix:sec:limitations_appendix}

Our systematic empirical analysis offers valuable insights into VLM-based OOD detection, certain inherent limitations define its scope and naturally point towards broader avenues for future research. Our findings are predominantly derived from specific, albeit representative, Vision-Language Model architectures (mainly CLIP variants); consequently, extending such empirical characterizations to the full diversity of rapidly evolving VLM paradigms (e.g., generative VLMs, larger backbones) and the entire spectrum of OOD problem types remains a continuous, expansive research endeavor.



\end{document}